\documentclass{article}
\usepackage{iclr2025_conference,times}

\usepackage{hyperref}
\usepackage{url}
\usepackage{graphicx}
\usepackage{xcolor}
\usepackage{colortbl}
\usepackage{booktabs}
\usepackage{amssymb}
\usepackage{pifont}
\usepackage{amsmath}
\usepackage{amsfonts}
\usepackage{amsthm}
\usepackage{tabularx}
\usepackage{tcolorbox}
\usepackage{pgfplots}
\usepackage{caption}
\usepackage{subcaption}
\usepackage{tikz}
\usepackage{pgfplotstable}
\usepackage{subcaption}
\usepgfplotslibrary{groupplots}
\pgfplotsset{compat=newest}

\theoremstyle{definition}

\graphicspath{ {./images/} }

\definecolor{myblue}{RGB}{70,130,180}
\definecolor{mygray}{RGB}{240,240,240}
\definecolor{darkblue}{RGB}{25,25,112}
\definecolor{lightblue}{RGB}{135,206,235}

\title{Gamified crowd-sourcing of high-quality data for visual fine-tuning}

\author{Shashank Yadav, Rohan Tomar, Garvit Jain, Chirag Ahooja, Shubham Chaudhary\\
Fraction AI\\
\texttt{\{shashank, rohan, garvit, chirag, shubham \}@fractionai.xyz} \\
\AND
Charles Elkan \\
Department of Computer Science \\
University of California, San Diego \\
\texttt{celkan@ucsd.edu} \\
}

\iclrfinalcopy % Uncomment for camera-ready version, but NOT for submission.
\begin{document}

\maketitle

\begin{figure}[h]
\centering
\subfloat(a){\includegraphics[width=0.3\textwidth]{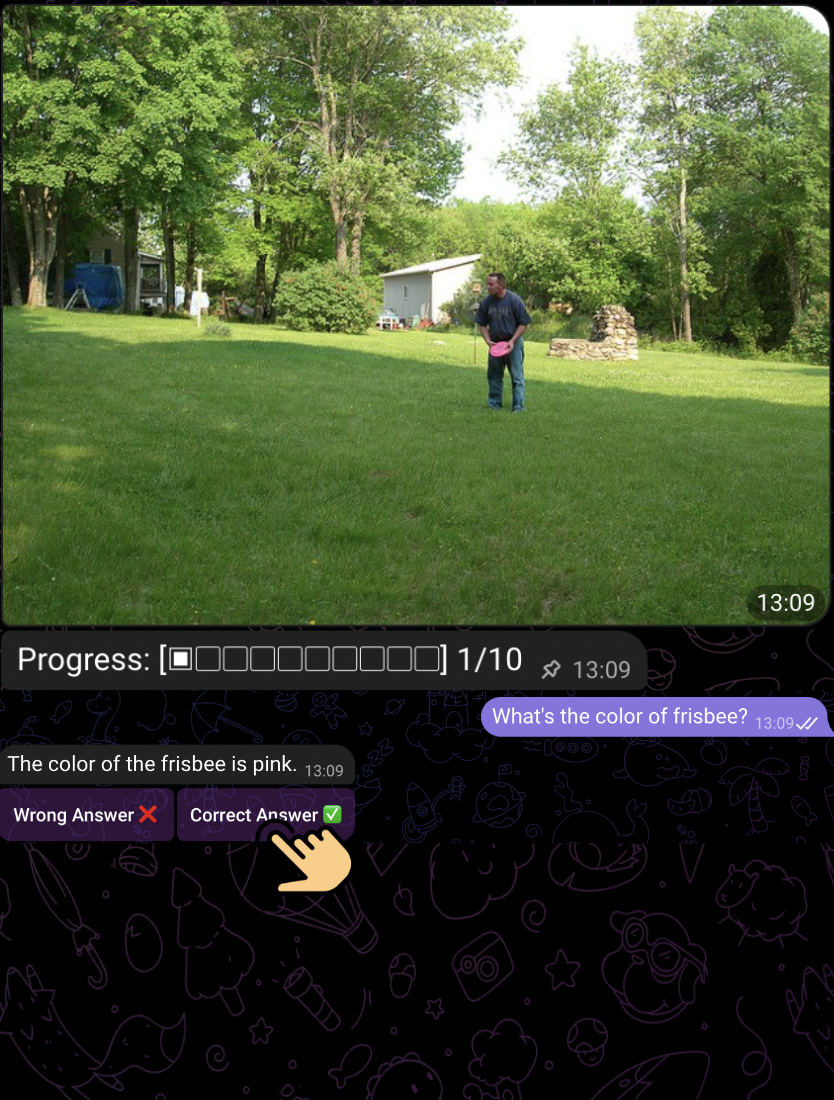}} 
\subfloat(b){\includegraphics[width=0.3\textwidth]{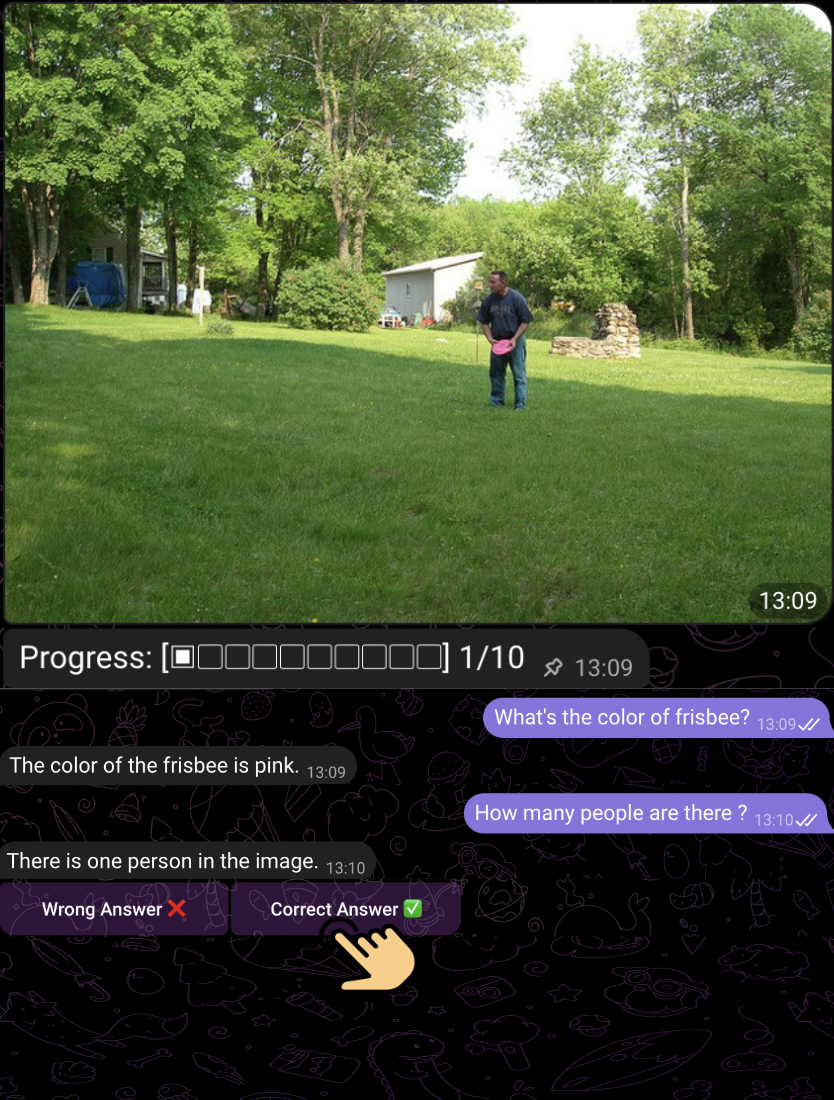}} 
\subfloat(c){\includegraphics[width=0.3\textwidth]{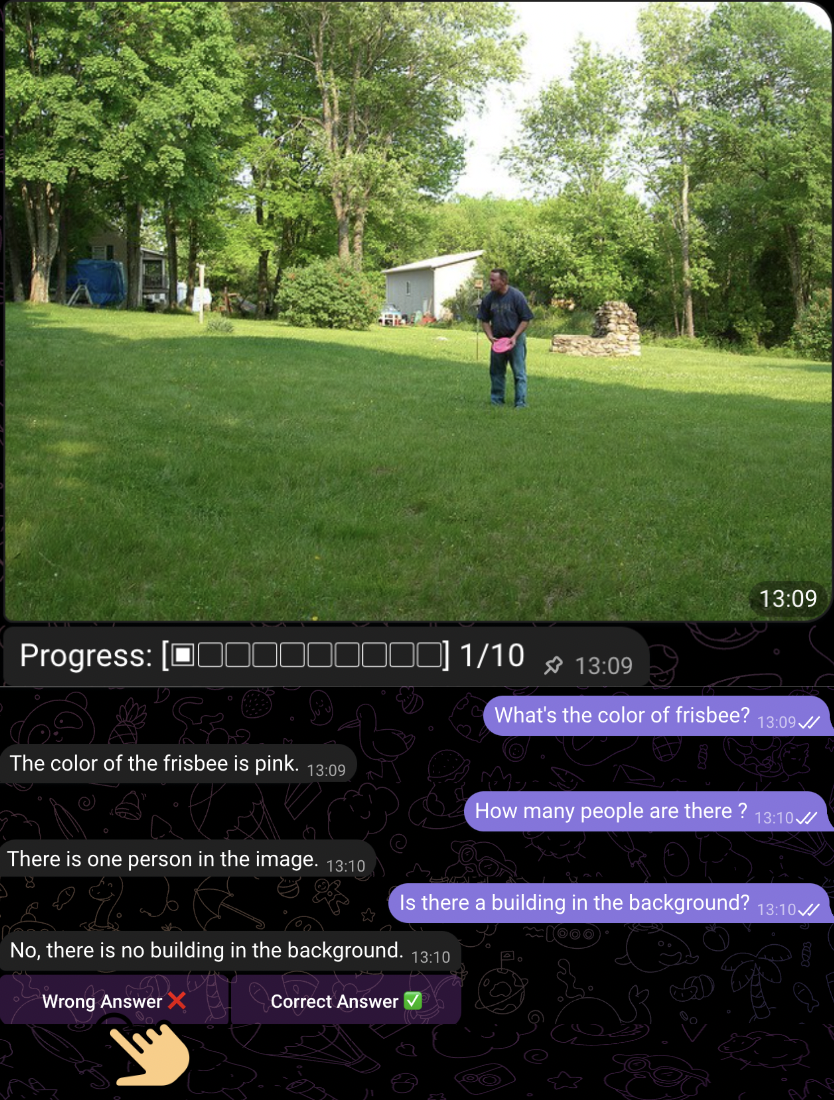}}
\caption{For each image, the goal of the player is to find a question that the AI answers incorrectly:
(a) player asks a question and the model answers correctly; 
(b) player tries again but the model answers right again; 
(c) finally, the model answers the third question wrong. 
There are 10 images in a session, 5 tainted and 5 untainted;
players don't know which images are tainted.
The player moves to the next image after 120 seconds or after choosing ``Wrong Answer." 
At the end of the session, they get 20 points per tainted image where they chose ``Wrong Answer"
and the model had been instructed to answer incorrectly.}
\label{fig:main_image}
\end{figure}

\begin{abstract}
This paper introduces {gamified adversarial prompting (GAP)}, 
a framework that crowd-sources high-quality data for visual instruction tuning of large multimodal models.
GAP transforms the data collection process into an engaging game, 
incentivizing players to provide fine-grained, challenging questions and answers that target gaps in the model's knowledge. 
% GAP enhances factual accuracy and reasoning, addressing the limitations of 
% current visual question answering (VQA) model capabilities and visual instruction tuning approaches.
Our contributions include 
(1) an approach to capture question-answer pairs from humans that directly address weaknesses in a model's knowledge,
(2) a method for evaluating and rewarding players that successfully incentivizes them to provide high-quality submissions,
and (3) a scalable, gamified platform that succeeds in collecting this data from over 50,000 participants in just a few weeks.
Our implementation of GAP has significantly improved the accuracy of a small multimodal model, namely MiniCPM-Llama3-V-2.5-8B, 
increasing its GPT score from 0.147 to 0.477 on our dataset, 
approaching the benchmark set by the much larger GPT-4V. 
Moreover, we demonstrate that the data generated using MiniCPM-Llama3-V-2.5-8B also enhances its performance across other benchmarks, and exhibits cross-model benefits. 
Specifically, the same data improves the performance of QWEN2-VL-2B and QWEN2-VL-7B
on the same multiple benchmarks.
%, highlighting the generalizability and transferability of the knowledge captured through GAP.
\end{abstract}

\section{Introduction}
Visual question answering (VQA) has emerged as a crucial paradigm in AI, 
extending beyond mere visual interpretation to facilitate broader and deeper understanding in models. 
Studies demonstrate VQA's potential in enhancing general knowledge acquisition, transfer learning, and complex reasoning skills. \cite{mahdisoltani2018effectivenesstaskgranularitytransfer} showed that pretraining on complex visual-linguistic tasks significantly improves performance across diverse downstream applications, from textual generation to fine-grained classification. 
The encoding of visual information as language, explored in works like Something-Else \citep{CVPR2020_SomethingElse, girdhar2019cater}, and more recently by \cite{alayrac2022flamingo}, enables models to develop low-level visual skills that support sophisticated reasoning in multimodal contexts. 
% Building on the grounded reasoning framework proposed by \cite{bhattacharyya2023look}, this paper examines how integrating VQA methodologies into AI training regimens can foster more versatile, intelligent systems capable of grounded reasoning across modalities, potentially bridging the gap between visual perception and higher-order cognition in artificial agents.

Large multimodal models (LMMs) have shown impressive capabilities in visual question answering tasks, demonstrating an ability to understand and reason about visual content \citep{liu2023visual, dai2023instructblip}. 
However, these models often struggle with fine-grained visual details, factual accuracy, and complex reasoning, 
particularly when faced with challenging, objective questions \citep{li2023evaluating, yu2023mm}. 
% These models undergo several stages of development, each serving a crucial role in shaping their performance:

% % \begin{figure}[b!]
% % \includegraphics[width=0.8\textwidth]{images/user_growth.png}
% % \centering
% % \caption{Weekly player growth on our Telegram miniapp.}
% % \label{fig:weekly_user_growth}
% % \end{figure}

% \begin{enumerate}
%     \item \textbf{Pretraining:} This stage builds a broad foundation of language and vision models.
%     \item \textbf{Supervised Fine Tuning (SFT):} This process adapts models to specific tasks or domains.
%     \item \textbf{Reinforcement Learning from Human Feedback (RLHF):} This stage aligns models with human preferences.
%     \item \textbf{Prompt engineering and few-shot learning:} These techniques optimize model usage for particular applications.
% \end{enumerate}

% Despite notable advancements across these processes, 
Supervised fine-tuning (SFT) remains a critical yet challenging phase for LLMs and LMMs,
especially in visual question answering.
SFT includes instruction tuning, where models are optimized to follow natural language instructions. 
Visual instruction tuning builds on this by fine-tuning models 
to handle both visual inputs and textual instructions such as questions or commands related to an image.
While visual instruction tuning has led to improvements in LMMs,
its effectiveness is constrained by the quality and specificity of the training data \citep{liu2023visual, li2023llava}.

% Recently introduced OpenAI-O1, based on Chain of Thought \cite{wei2022chain}, is supposedly trained by generating series of steps to solve a problem. These steps and the final output are judged by an external LLM. Objective correctness of the external LLM is necessary for this process to work.\\ 

% To address these challenges, we propose Gamified Adversarial Prompting (GAP), a novel framework for enhancing the capabilities of LMMs in VQA tasks. GAP focuses on gathering question answer pairs related object types, counting objects, object actions, object locations, the relative positions between objects, classifying actions, describing actions, and other related details from images.

Our gamified adversarial prompting (GAP) method transforms visual instruction tuning
by incentivizing players to come up with challenging, fine-grained questions that target specific weaknesses in a given LMM.
The GAP approach focuses on improving factual accuracy and reasoning capabilities, 
particularly for objective, verifiable information more than for subjective preferences or general alignment. 

% By gamifying the process of identifying and addressing LMM weaknesses, GAP creates a dataset that pushes the boundaries of model capabilities, addressing the limitations of current visual instruction tuning approaches \cite{fu2023mme, liu2023mmbench}.

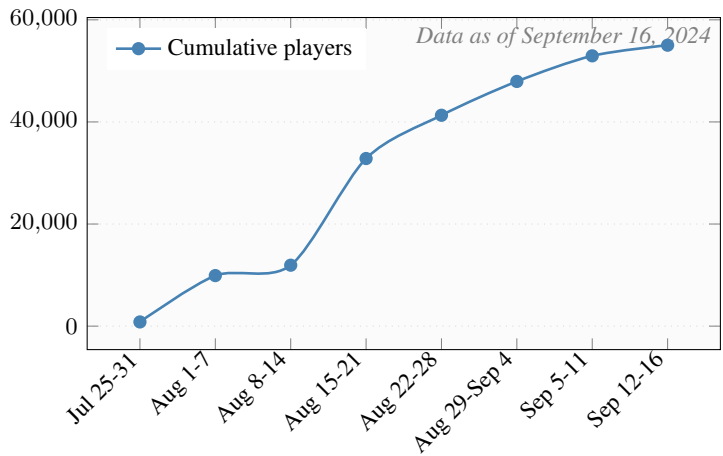
\begin{figure}[t]
\centering
\begin{tikzpicture}
\begin{axis}[
    width=10cm,
    height=6cm,
    ylabel style={font=\small\bfseries},
    xticklabels={
        {Jul 25-31},
        {Aug 1-7},
        {Aug 8-14},
        {Aug 15-21},
        {Aug 22-28},
        {Aug 29-Sep 4},
        {Sep 5-11},
        {Sep 12-16}
    },
    xtick=data,
    x tick label style={rotate=45,anchor=east,font=\footnotesize},
    ymajorgrids=true,
    grid style={dotted,color=gray!30},
    ylabel near ticks,
    scaled y ticks=false,
    yticklabel style={/pgf/number format/fixed, /pgf/number format/precision=0, font=\footnotesize},
    axis background/.style={fill=mygray!30},
    legend pos=north west,
    legend style={font=\footnotesize, fill=white, draw=none},
    tick label style={font=\footnotesize},
]

\addplot[
    color=myblue,
    mark=*,
    line width=1pt,
    mark size=2pt,
    smooth,
] coordinates {
    (1,821)
    (2,9899)
    (3,11935)
    (4,32832)
    (5,41305)
    (6,47929)
    (7,52947)
    (8,55030)
};
\addlegendentry{Cumulative players}

\node[anchor=north east, font=\footnotesize\itshape, color=gray] at (rel axis cs:1,1) {Data as of September 16, 2024};

\end{axis}
\end{tikzpicture}
\caption{Weekly player growth on our Telegram miniapp.}
\label{fig:weekly_user_growth}
\end{figure}

The contributions of this paper include an approach to automatically evaluate and reward player submissions with high accuracy 
inspired by the reCAPTCHA approach \citep{ahn2008}. 
This evaluation scheme ensured we were able to scale GAP to over 50,000 players (Figure \ref{fig:weekly_user_growth}) 
in just a few weeks, while ensuring high data quality and player engagement.

% In the following sections, we detail the GAP framework, our data collection process, and our evaluation methodology. We then present our results, demonstrating the effectiveness of datasets generated through GAP in improving LMM performance on challenging VQA tasks. Finally, we discuss the implications of our work and potential future directions for research in this area.

\section{Related Work}

\paragraph{Visual Question Answering (VQA)}

Recent advancements in Visual Question Answering (VQA) have focused on improving model architectures and training methodologies. Our work builds upon these foundations while introducing a novel approach to data collection and model finetuning.

InstructBLIP \citep{dai2023instructblip} adopted Q-Former to sample from visual tokens for LLM processing. 
LLaVA \citep{liu2023visual} introduced visual instruction tuning, using instruction-following data to convert LLMs into multimodal LLMs. 
Flamingo \citep{alayrac2022flamingo} and IDEFICS \citep{laurencon2024matters} used shared decoders for visual-language understanding. 
Qwen-VL \citep{bai2023qwen} employed a three-stage training process to convert LLMs to multimodal models. 
SPHINX \citep{gao2024sphinx} adopted multiple visual encoders to enrich visual features. 
InternLM-Xcomposer \citep{zhang2023internlm} achieved state-of-the-art performance using interleaved text-image composition data. 

Recently, \cite{yao2024minicpmv} introduced MiniCPM-V, a series of efficient LMMs designed for end-side devices. Their work demonstrates the potential for GPT-4V level performance with significantly fewer parameters, while being deployable on mobile phones. This represents a trend towards more efficient and accessible LMMs. We use instruction tuned MiniCPM-Llama3-V-2.5-8B as our base model owing to the small size coupled with high performance.

On other recent advancements in LMMs, \cite{lu2024deepseek} aim for real-world vision-language understanding, 
while \cite{li2024llava} improve reasoning, OCR, and world knowledge in LMMs. 
CuMo \citep{li2024cumo} introduced a method for scaling LMMs using mixtures of experts (MoE), 
incorporating sparse MoE blocks into both the vision encoder and the MLP connector. 
Recently, \cite{hong2024cogvlm2} proposed CogVLM2, a family of visual language models that achieve state-of-the-art performance on various image and video understanding benchmarks.

\paragraph{Adversarial and Challenging Dataset Creation}

Creating challenging datasets that expose AI system limitations has been a focus of recent research. 
SWAG \citep{zellers2018swag}, HellaSwag \citep{zellers2019hellaswag}, and Social IQa \citep{sap2019social} introduced large-scale adversarial datasets for commonsense inference. 
WinoGrande \citep{sakaguchi2020winogrande} presented an adversarial dataset for commonsense reasoning at scale.

Researchers have also explored algorithmic methods for bias reduction, such as adversarial filtering \citep{zellers2018swag,zellers2019hellaswag,le2020adversarial}. 
These approaches help mitigate bias and reduce dataset artifacts, though they are often applied post-hoc. 
Our vast and geographically diverse contributor base potentially yields less biased data by incorporating a wide range of global perspectives.

MME \citep{fu2023mme} contains manually annotated instruction-answer pairs to measure perception and cognition abilities on a total of 14 subtasks. 
DocVQA \cite{mathew2021docvqa}, another large dataset, consists of 50,000 questions defined on 12,000+ document images. \cite{liu2023mmbench} incorporate multiple-choice questions in both English and Chinese versions.
In the current work, we focusing only on English.
These datasets been created by crowd-sourcing. 
It is difficult to scale this approach while maintaining quality;
our approach is an improvement on this front.

\paragraph{Benchmarks}
The most recent benchmarks include OCRBench \citep{liu2024hiddenmysteryocrlarge} designed to assess optical character recognition (OCR) capabilities; 
MMMU \citep{yue2023mmmu} consisting of multimodal questions from college exams, quizzes, and textbooks spanning 30 subjects and 183 subfields; 
MM-Vet \citep{yu2024mm} is another benchmark that assesses LMMs on complex tasks requiring the integration of multiple vision-language capabilities; 
HallusionBench \citep{guan2023hallusionbench}, challenging benchmark targeted towards LMM hallucinations. 
All these are available through VLMEvalKit \citep{duan2024vlmevalkit}
We use these benchmarks to evaluate the model finetuned on the dataset generated through GAP.

\paragraph{Gamification and Human-in-the-Loop Approaches}

Our gamified approach to data collection is inspired by works leveraging human interaction to create challenging datasets. 
CommonsenseQA 2.0 \citep{talmor2021commonsenseqa} introduced a gamification framework where players compose questions to mislead a rival AI. 
This approach led to enhanced player engagement and high-quality data collection at scale.

Other relevant human-in-the-loop approaches include Beat-the-AI \citep{bartolo2020beat}, which investigated adversarial human annotation for reading comprehension, 
StrategyQA \citep{geva2021did}, which created a benchmark with implicit reasoning strategies, 
and Dynabench \citep{kiela2021dynabench}, which proposed dynamic dataset creation. 
These works demonstrate the value of incorporating human intelligence into the training process.

Fool Me Twice \citep{eisenschlos2021fool} introduced a game for collecting entailment data through multi-player interaction, 
further illustrating the potential of gamification in dataset creation.

Tools like \cite{shnarch2022labelsleuth}, \cite{labelstudio} and \cite{kim2024megannohumanllmcollaborativeannotation} while not gamified, 
improve human labelling by using a pretrained AI model. 

\paragraph{Large Scale Data Collection}
\cite{laurencon2024matters} has released a massive collection of 50 vision-language datasets covering a wide range of tasks: 
general visual question answering, counting, captioning, text transcription, document understanding, chart/figure understanding, table understanding, visual reasoning,
geometry, spotting differences between 2 images or converting a screenshot to a functional code. 
Our base model MiniCPM-Llama3-V-2.5-8B \citep{yao2024minicpmv} has been trained on this dataset.

In contrast to \cite{chen2023sharegpt4v} and \cite{laion2023gpt4v} who collected detailed image caption data from GPT-4V to augment VQA models, 
we use a gamified human in the loop approach, 
maintaining legal and ethical compliance while enhancing transparency in model improvement. 
This strategy builds our model on a foundation of original, human-verified data, avoiding potential issues associated with AI-generated content.
We make part of the data generated during the study publicly available for future research.

\paragraph{Model Analysis and Robustness}

Recent work has highlighted the importance of thorough model analysis. 
Studies have shown that while models achieve human parity on many benchmarks, they can be brittle when faced with out-of-domain \citep{brown2020language,talmor2019multiqa,fisch2019mrqa} or adversarial examples   \citep{jia2017adversarial,rajpurkar2018know,yuan2019adversarial,ribeiro2018semantically}. 
\cite{yuan2023revisiting, raina2024llm} observed these issues in LLMs as well.

Contrast sets \citep{gardner2020evaluating} and counterfactual data \citep{kaushik2020learning} have been used to examine model consistency, revealing limitations in current state-of-the-art models. 
Our dataset generated through GAP can be visualized as a contrast set in relation to the pre-existing knowledge base of an LMM.

% By combining elements of gamification, human-in-the-loop learning, efficient data collection, and targeted finetuning, our work aims to address the challenges of creating more robust models while engaging players in the process.

\section{Gamified Adversarial Prompting (GAP)}
% In this section we talk about the kind of data we believe would be useful for visual finetuning. Then we talk about game design principles that incentivizes player behaviour towards generating that kind of data. Finally we talk about how we're able to evaluate the submissions automatically at scale

% \paragraph{Adversarial Questions and Informative Samples}
When it comes to data, quality beats quantity \citep{zhou2024lima, li2023quantity}. 
% We come up with a plausible way to define what kind of datapoints could be useful for improving model performance. 
Given an LMM $M$, an image $i$ and question $q$ related to the image and the expected answer $a$, 
the model response is denoted $r = M(i, q)$.
% \begin{definition}[Adversarial Question]\label{adversarial_ques}
% The $q$ is defined as an adversarial question wrt $M$ and $i$ if the response $r$ is not factually equivalent to the expected answer $a$.
% \end{definition}
% \begin{definition}[Informative Sample]\label{informative_sample}
The tuple $(i, q, a)$ is defined as an \textbf{informative sample} for $M$ if the response $r$ is not factually equivalent to the expected answer $a$.
% \end{definition}

While LMMs are stochastic and generate different responses to the same query, 
generating a factually incorrect response is usually a sign of a gap in a model's knowledge. 
% We design our game towards generating Informative samples hence specifically targeting the gaps in model knowledge. We further evaluate how usefulness of informative samples in improving model performance over different benchmarks.

\paragraph{Gameplay}
One session for one player proceeds as follows:
\begin{enumerate}
\item Player sees an image
\item They ask a question which they believe AI would answer incorrectly
\item Player evaluates the AI's response, they have to mark it "Correct" or `Wrong`
\item If they mark it `Wrong` they move to the next image, otherwise they can ask more questions. Player automatically moves to the next image after 120 secs.
\end{enumerate}
For every session, a player sees 10 images and has at most 120 seconds for each image. 
When the player finishes 10 images, or the session resets after 6 hours, they earn points based on their performance.
% At the end of the week, top 3 participants based on points earn cash rewards of \$50, \$30 and \$20 for first, second and third positions respectively.
We impose a time limit on each image to create a sense of urgency. Limiting sessions to 10 images keeps casual players motivated by allowing them to engage briefly while still competing for rewards

\paragraph{Image Datasets}

\begin{figure}[h]
\centering
(a){\includegraphics[width=0.45\textwidth]{images/tainted_image.jpg}} 
(b){\includegraphics[width=0.45\textwidth]{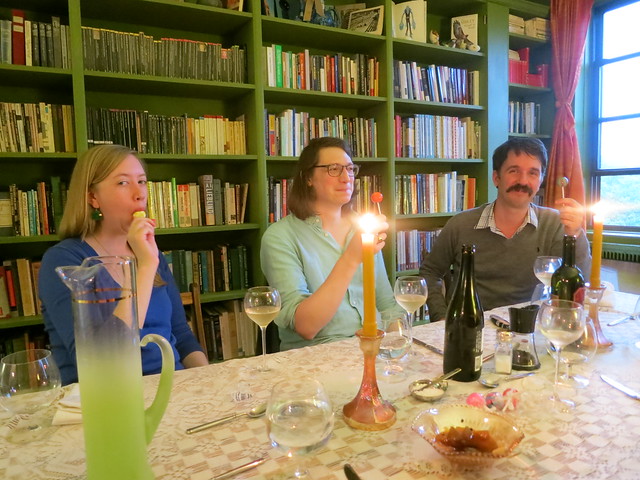}} 
\caption{(a) Sample image from the tainted dataset, a few easily identifiable items with basic lighting; (b) Sample from the untainted dataset, multiple people and items with complex lighting.}
\label{fig:tainted_vs_untainted}
\end{figure}

The GAP framework utilizes two sets of image datasets derived from MS-COCO \cite{lin2014microsoft}:
\begin{itemize}
    \item The \textbf{tainted dataset} initially comprises 1000 carefully curated images. The creation process involves selecting images with two or fewer objects, generating detailed descriptions using our Base Model (MiniCPM-Llama3-V-2.5-8B), and manually reviewing to select images with complete and error-free descriptions. For these images, the AI model is expected to have complete knowledge.

    \item The \textbf{untainted dataset} includes the remaining, more complex COCO images, used to challenge the LMM.%
\footnote{
It could be the case that players learn (maybe unconsciously) the pattern that distinguishes tainted from untainted images. They could then learn that what they do with untainted images does not affect their scores, and then provide bogus input for untainted images. Players will do this if they are motivated by speed, even if they are not hostile. In practice the concern has not materialized as observed with manual evaluations on subsets of data over time. Moreover, after we obtain more than 10 informative data points for an image from the untainted set and the model is fine-tuned on that data, the said image is moved to the tainted set. This helps grow the tainted set and reduces the differences between the two sets.
}
\end{itemize}
% This splitting of datasets ensures we have a Tainted set where our model is expected to be correct all the time. So we can artificially manipulate them to show players incorrect responses and use that for evaluation.

\paragraph{Evaluation and Scoring}
\begin{table}[t]
\centering
\begin{tabular}{|c|c|c|c|}
\hline
\rowcolor{gray!20}
\textbf{Dataset (D)} & \textbf{Asked for Incorrect (I)} & \textbf{Model Correct (M)} & \textbf{Human Marked Correct (H)} \\
\hline
Tainted & \checkmark & \checkmark & \checkmark \\
\rowcolor{gray!20}
Tainted & \checkmark & \checkmark & \ding{55} \\
Tainted & \checkmark & \ding{55} & \checkmark \\
\rowcolor{gray!20}
Tainted & \checkmark & \ding{55} & \ding{55} \\
Tainted & \ding{55} & \checkmark & \checkmark \\
\rowcolor{gray!20}
Tainted & \ding{55} & \checkmark & \ding{55} \\
Tainted & \ding{55} & \ding{55} & \checkmark \\
\rowcolor{gray!20}
Tainted & \ding{55} & \ding{55} & \ding{55} \\
\hline
\multicolumn{4}{|c|}{\cellcolor{gray!10}\textit{}} \\
\hline
Untainted & \ding{55} & \checkmark & \checkmark \\
\rowcolor{gray!20}
Untainted & \ding{55} & \checkmark & \ding{55} \\
Untainted & \ding{55} & \ding{55} & \checkmark \\
\rowcolor{gray!20}
Untainted & \ding{55} & \ding{55} & \ding{55} \\
\hline
\end{tabular}
\caption{All possible cases for player interaction: which dataset,
whether we asked for an incorrect response,
whether the response the player saw was correct, 
and whether the player marked the response as correct.}
\label{tab:conditional_prob}
\end{table}

In each session a player interacts with five images from the tainted dataset and five from the untainted one. 
Users are unaware of this split. 
For each question asked concerning an image from the tainted set, 
we randomly instruct the model to change its answer subtly, 
introducing a minor inaccuracy. 
This process ensures that the incorrect answers are plausible and closely related to the truth, making the challenge more nuanced.
% During gameplay, the AI randomly presents either the correct answer or the slightly perturbed incorrect version. 
Participants are scored based on their ability to identify these deliberate mistakes.
% providing a reliable measure of their critical thinking and attention to detail. 
% This method allows us to control the difficulty level and maintain a consistent evaluation metric across different images and questions

% There are several possibilities based on Table \ref{tab:conditional_prob}:
% \begin{enumerate}
%     \item If the image comes from Tainted or Untainted dataset
%     \item If we explicitly asked the model for an Incorrect response
%     \item If the model response is correct
%     \item If the player marks the model response as correct
% \end{enumerate}
% Based on our definition of Adversarial Questions \ref{adversarial_ques}, 
Our goal is to obtain questions on the untainted dataset where the model answers incorrectly. 
We want to encourage players to ask difficult questions where the model gives an incorrect response,
but we also want to disincentivize players from marking correct model responses as incorrect. So our goal is twofold:
(1) evaluate which questions from the untainted dataset are incorrectly answered by the model;
(2) reward the players appropriately to encourage only desired behavior.

% \paragraph{Evaluation}
Let $D$, $I$, and $M$ be random variables defined as in Table \ref{tab:conditional_prob}. 
Let
\begin{align}
P(M = 1 \mid D &= \text{Tainted}, I = 0) = 1 - \varepsilon \label{eq:correct_not_asked} \\
P(M = 0 \mid D &= \text{Tainted}, I = 0) = \varepsilon \label{eq:incorrect_not_asked} \\
P(M = 1 \mid D &= \text{Tainted}, I = 1) = \delta \label{eq:correct_asked} \\
P(M = 0 \mid D &= \text{Tainted}, I = 1) = 1 - \delta \label{eq:incorrect_asked}
\end{align}
where $\varepsilon$ and $\delta$ are small positive numbers such that $\delta < \varepsilon$.

The tainted dataset is constructed to be simpler and more straightforward, as mentioned previously. 
When not explicitly asked for an incorrect answer ($I = 0$), 
the model's high accuracy on the tainted dataset is represented by Equation \ref{eq:correct_not_asked}. 
The $\varepsilon$ term accounts for the possibility of errors due to random fluctuations or edge cases.
Conversely, the probability of the model answering incorrectly on the tainted dataset when not asked to do so is low
(Equation \ref{eq:incorrect_not_asked}).
    
However, when asked for an incorrect answer ($I = 1$), the model is likely to provide one (Equation \ref{eq:incorrect_asked}).
% This high probability reflects the model's capability to generate plausible incorrect answers when instructed, even for simple images.
There is a small chance the model answers correctly when asked for an incorrect answer (Equation \ref{eq:correct_asked}).
The relationship $\delta < \varepsilon$ indicates that the model is more likely to give an incorrect answer when asked,
than to give a correct answer when not asked on the tainted dataset. 
This reflects the strong influence of the explicit instruction to provide an incorrect answer, 
overriding the model's natural tendency towards accuracy on the simplified tainted dataset.
% This characteristic makes the tainted dataset useful for controlled studies of model behavior 
% and for generating specific types of model outputs. 
% It allows us to reliably manipulate the model's responses while maintaining a high degree of predictability in its performance.

Assume that $P(M=0 \mid H=0, \text{Untainted}) = P(M=0 \mid H=0, \text{Tainted}).$
% Given $\varepsilon, \delta \to 0$, 
Using Bayes' rule,
\begin{equation}
    P(M=0 \mid H=0, D=\text{Tainted}) = \frac{P(H=0 \mid M=0, D=\text{Tainted}) \cdot P(M=0 \mid D=\text{Tainted})}{P(H=0 \mid D=\text{Tainted})}
\end{equation}
For small $\varepsilon$ and $\delta$, % \to 0$, 
\begin{equation}\label{eq:approximation}
    P(M=0 \mid D=\text{Tainted}) \approx P(I=1 \mid D=\text{Tainted}).
\end{equation}
For the denominator, we use the law of total probability, dropping D $=$ Tainted for brevity in the right hand side:
\begin{equation}
    P(H=0 \mid D=\text{Tainted}) = P(H=0 \mid I=0) \cdot P(I=0) + P(H=0 \mid I=1) \cdot P(I=1)
\end{equation}
Combining these results,
\begin{equation}
    P(M=0 \mid H=0, D=\text{Untainted}) \approx \frac{P(H=0 \mid I=1) \cdot P(I=1)}{P(H=0 \mid I=0) \cdot P(I=0) + P(H=0 \mid I=1) \cdot P(I=1)}
\end{equation}
We update this measure at every Tainted sample and choose to select a question as an adversarial question 
whenever $P (M = 0 | H = 0, D = \text{Untainted}) > \theta$. 
In our experiments $\theta = 0.8$ gives good results.\label{sample_selection}

\paragraph{Reward system}
On a tainted image, the player earns 20 points every time 
they mark a model response incorrect ($H=0$) and it is actually incorrect ($M=0$).
No reward is given on untainted images since we cannot evaluate those. 
The expected reward is
\begin{equation}
R = \sum_{n=1}^{5}20 \cdot P(M=0 \mid H=0, D = \text{Tainted}) \sim \sum_{n=1}^{5}20 \cdot P(I=1 \mid H=0, D = \text{Tainted})\\    
\end{equation}
using \ref{eq:approximation} for approximation.

% \paragraph{Reward System}
To maintain high levels of engagement and ensure data quality, 
GAP incorporates a multi-faceted reward system. Self-Determination Theory (SDT) \citep{deci2012self} identifies two types of motivation:
intrinsic and extrinsic. Intrinsic motivation involves doing something for inherent enjoyment or interest, 
like playing a video game for fun. 
Extrinsic motivation is driven by external outcomes, such as studying for good grades. 
The quality of experience and performance can differ significantly between intrinsically and extrinsically motivated behaviors
\citep{kiela2021dynabench, cheah2022motivations, kanellopoulou2020engage}.
Intrinsic motivation leads to better results in the longer term but extrinsic motivation has significant positive impact also.

The game incorporates mechanics to drive player motivation through intrinsic and extrinsic factors:
\begin{enumerate}
    \item \textbf{Experience Points (XP):} Players earn 20 points for each correct answer the AI misses on a tainted dataset, providing immediate feedback.
    \begin{itemize}
        \item \textit{Extrinsic:} Points act as an external reward.
        \item \textit{Intrinsic:} Points foster a sense of competence and mastery.
    \end{itemize}

    \item \textbf{Leaderboard:} A weekly leaderboard highlights top performers based on points.
    \begin{itemize}
        \item \textit{Extrinsic:} Public recognition motivates.
        \item \textit{Intrinsic:} Competition satisfies the need for relatedness.
    \end{itemize}

    \item \textbf{Web3 Airdrops:} Future cryptocurrency rewards incentivize high-quality contributions.
    \begin{itemize}
        \item \textit{Extrinsic:} Clear external incentives.
        \item \textit{Intrinsic:} Appeal to players interested in Web3 innovation.
    \end{itemize}

    \item \textbf{Cash Prizes:} Top-ranking players receive \$150 in cash, with rewards for both the top 3 and 3 random players from the top 10.
    \begin{itemize}
        \item \textit{Extrinsic:} Pure cash incentives.
        \item \textit{Intrinsic:} Random prizes create hope and excitement.
    \end{itemize}
\end{enumerate}

This system encourages both short-term engagement and long-term commitment, aiming to gradually transition to token-based rewards with in-game mechanics like levels, NFTs, and guilds. 
While crypto-native players initially prefer cash and token rewards, leaderboard ranking also drives engagement, even without immediate financial gains. 
% As the game evolves, more engaging elements will improve retention.

\section{Experimental Setup}
We launched our platform as a Telegram miniapp, engaging over 50,000 participants within weeks, without traditional marketing. 
This rapid growth highlights Telegram's effectiveness for player acquisition in the crypto space. 
The success of {tap-to-earn} games like \textit{Catizen} (30 million players) and \textit{Hamster Kombat} (300 million players) further demonstrates the platform's potential. 
Telegram was chosen over Discord for its larger crypto-native user base and stronger viral growth potential for blockchain projects.
% Future iterations will focus on strategies to increase user participation and improve data quality.

\paragraph{Baseline Model}
We selected MiniCPM-Llama3-V-2.5-8B as our baseline model due to its state-of-the-art performance in multimodal tasks, cost-efficiency, scalability, and flexible Apache 2.0 licensing. The model demonstrates comparable capabilities to LLaVA \citep{liu2023visual}, Phi-3 \citep{abdin2024phi}, and Qwen-VL \citep{bai2023qwen, wang2024qwen2}, particularly in single-image understanding and visual question answering. Table \ref{tab:model-performance} summarizes the key characteristics of this model.

\begin{table}[t]
\centering
\begin{tabular}{ll}
\hline
Metric & Value \\
\hline
Inference speed & 75 input text tokens/second \\
Memory footprint & 19 GB GPU memory \\
Throughput & 0.3 requests/second \\
Average request & 250 input tokens, 640x428 pixel image \\
Cost per 1000 tokens & \$0.0037 \\
Hardware & AWS g5.xlarge (1 A10G tensor chip, 24GB memory) \\
\hline
\end{tabular}
\caption{Performance characteristics of MiniCPM-Llama3-V-2.5-8B}
\label{tab:model-performance}
\end{table}

The model's efficient resource utilization enables effective horizontal scaling, making it suitable for large-scale experiments and potential real-world applications. 
These characteristics provide a strong foundation for evaluating improvements in multimodal language understanding and generation tasks while considering practical deployment constraints.

\paragraph{GAP-VQA Dataset and Experiment Design}
To create {GAP-VQA}, we randomly sampled $3,683$ question-image pairs, filtering them using a threshold $\theta = 0.8$ (Equation~\ref{sample_selection}) to ensure a high proportion of adversarial examples. 
While this filtering may include some non-adversarial questions, these do not negatively impact the model; 
instead, they contribute to the overall diversity of the dataset while having a less pronounced benefit.

Our labeling process combines human expertise with AI refinement to produce high-quality, accurate answers for these visual questions. Initially, trained in-house labeling experts provide answers based on the given images, ensuring a high baseline of accuracy and leveraging human understanding of context and nuance. Subsequently, we employ Llama 3.1 8B \citep{dubey2024llama} to refine these answers, correcting errors, improving clarity, and maintaining consistency across the dataset.

% This two-step approach harnesses both human insight and advanced language model capabilities, resulting in a linguistically robust dataset that evolves with improvements in open-source models. The process maintains full control over our pipeline, crucial for effectively training and evaluating multimodal AI systems.

GAP-VQA is divided into two subsets: {GAP-VQA-train}, consisting of 2,683 question-answer pairs, and {GAP-VQA-val}, comprising 1,000 question-answer pairs. 
% This division facilitates rigorous model evaluation and assessment of our iterative fine-tuning process, enabling comprehensive analysis of model performance on diverse, adversarial visual questions generated through GAP.
% \paragraph{Experimentation and Evaluation}
% Our experimentation process is designed to rigorously evaluate the data collected through our gamified platform. We aggregate all the data generated during the continuous improvement cycle and divide it into train, validation, and test sets for structured analysis. 
We carry out three separate types of evaluations:

\begin{table}[h!]
\centering
\begin{tabularx}{\textwidth}{|c|X|}
\hline
\textbf{Experiment Type} & \textbf{Description} \\
\hline
\text{Same dataset} & Fine-tune MiniCPM-Llama3-V-2.5-8B on  GAP-VQA-train, then assess accuracy on GAP-VQA-val. This provides a direct measure of how well the model learns to generalize across the GAP-VQA dataset.\\
\hline
\text{Cross dataset}  & Evaluate the fine-tuned model on other benchmarks, such as MME, OCRBench, MMBench, and TextVQA. This tests how improved accuracy transfer to standardized multimodal tasks.\\
\hline
\text{Cross model}    & Fine-tune QWEN2-VL-2B and QWEN2-VL-7B on GAP-VQA-train.
%(Note that GAP-VQA-train has been generated using MiniCPM-Llama3-V-2.5-8B as base model). 
Evaluate these models on GAP-VQA-val and other benchmarks. 
This determines whether the GAP dataset for one model benefits another model, indicating its potential for broader AI advancement.\\
\hline
\end{tabularx}
\caption{Overview of experiment types}
\end{table}

All models are fine-tuned using LoRA \citep{hu2022lora} through Llama Factory \citep{zheng2024llamafactory}. 
We opt for LoRA over full fine-tuning to reduce GPU memory usage and train fewer parameters, enabling faster experimentation and lower computational costs. LoRA also mitigates catastrophic forgetting and improves generalization by preserving pre-trained knowledge

\section{Results}
% The GAP-VQA-train dataset contains diverse set of questions and can be divided into 13 different groups, which are listed in Table~\ref{tab:vqa-categories} in the Appendix. The questions test many ways of understanding images, from easy to hard. Some questions are simple, like naming objects or describing what's in a picture. Others are harder, such as finding odd things or explaining cultural events shown in images. These questions check different skills. They ask about naming objects and describing their features, understanding where things are in a picture, and counting things. Some questions involve reading words in images, recognizing feelings and actions, or guessing the time of day or season. Others focus on describing weather and surroundings or understanding art styles. This wide range of questions help LMMs generalize better.

The GAP-VQA-train dataset includes a diverse set of questions, divided into 13 groups (Table~\ref{tab:vqa-categories} in the Appendix). 
These questions assess varying levels of image understanding, 
from simple tasks like object naming and description to more complex ones like identifying anomalies or explaining cultural events. 
They evaluate object recognition, spatial reasoning, counting, reading text in images, recognizing emotions, estimating time or season, and interpreting art styles. 
This broad range of questions helps LMMs improve their generalization capabilities.

\paragraph{Model Performance on the GAP-VQA-val Dataset}

We evaluated the performance of various models on the GAP-VQA-val dataset using GPT-4 as an evaluator. 
A comprehensive prompt was used to assess the model's answers based on criteria including existence, position, count, and color accuracy,
with scores ranging from 0 to 1, where 1 indicates perfect alignment.
The evaluation prompt is detailed in Appendix~\ref{appendix:gpt-4v_evaluation}.

% The GAP-VQA dataset was specifically designed by creating questions that challenged the MiniCPM model, identifying areas where the model struggled. This guided the fine-tuning process by emphasizing areas where MiniCPM showed poor performance, leading to better alignment during training. As a result \ref{tab:gap-vqa-performance}, MiniCPM-Llama3-V-2.5-8B exhibited the largest improvement post-fine-tuning, with a score increase of +0.300, from 0.147 to 0.447. This substantial gain indicates that the dataset allowed MiniCPM to better capture challenging multimodal relationships after fine-tuning.

% In comparison, Qwen2-VL-2B and Qwen2-VL-7B showed more modest improvements of +0.116 and +0.043, respectively. This smaller gain suggests that Qwen2’s architecture was less effective at capturing the dataset's complexity during fine-tuning. Potential reasons include the model’s smaller size and less aligned pre-trained representations for the types of visual reasoning required. Addressing these limitations with more extensive pre-training or tailored fine-tuning techniques could help improve Qwen2's adaptability in future iterations.

The GAP-VQA dataset was designed to challenge the MiniCPM model by targeting its weaknesses, 
guiding the fine-tuning process for better alignment. 
As shown in Table~\ref{tab:gap-vqa-performance}, MiniCPM-Llama3-V-2.5-8B achieved major improvement post-fine-tuning, with a score increase of +0.300 (from 0.147 to 0.447). 
This indicates the dataset's effectiveness in helping capture complex multimodal relationships.
In comparison, Qwen2-VL-2B and Qwen2-VL-7B saw more modest improvements of +0.116 and +0.043, respectively, 
suggesting Qwen2's architecture was less suited for the dataset's complexity. 
The smaller size of these models, and less aligned pre-training, likely contributed to the weaker improvements.
%which could be addressed with more extensive pre-training or tailored fine-tuning.

\begin{table}[tbp]
\centering
\begin{tabular}{lccc}
\toprule
Model & Pre-fine-tuning & Post-fine-tuning & Improvement \\
& GPT Score & GPT Score & \\
\midrule
GPT-4V (Benchmark) & 0.637 & - & - \\
MiniCPM-Llama3-V-2.5-8B & 0.147 & 0.477 & +0.300 \\
Qwen2-VL-2B & 0.169 & 0.285 & +0.116 \\
Qwen2-VL-7B & 0.207 & 0.250 & +0.043 \\
\bottomrule
\end{tabular}
\caption{Performance on the GAP-VQA-val dataset 
before and after fine-tuning. 
GPT-4V is a benchmark and is not fine-tuned.}
\label{tab:gap-vqa-performance}
\end{table}

% Overall, while all models benefited from fine-tuning, 
% the dataset's design around MiniCPM's weaknesses resulted in significantly better improvements for MiniCPM compared to the Qwen2 models.

\paragraph{Cross Dataset Evaluation}
Our GAP-VQA approach demonstrates that targeted knowledge gap filling significantly enhances model generalization (Table~\ref{tab:performance_comparison_updated}). 
By training on carefully curated data based on COCO \citep{lin2014microsoft}, 
the MiniCPM-Llama3-V-2.5-8B model exhibits improved performance across diverse benchmarks, including OCR, hallucination detection, and complex reasoning tasks.
This cross-dataset improvement suggests that addressing specific knowledge deficits enables models to develop more robust and transferable skills. 
The enhanced performance on benchmarks like MME, TextVQA, and MM-Vet, whose contents differ substantially from the training data, underscores the effectiveness of our method in promoting generalization.
% These findings highlight the potential of strategic dataset curation in advancing AI capabilities across multiple domains, offering a promising direction for developing more versatile and adaptable models.

\begin{table}[tbp]
\centering
\small
\begin{tabular}{lrr}
\toprule
Dataset & \multicolumn{1}{c}{MiniCPM-Llama3-V-2.5-8B} & \multicolumn{1}{c}{MiniCPM-Llama3-V-2.5-8B} \\
& \multicolumn{1}{c}{(orig)} & \multicolumn{1}{c}{(finetuned on GAP-VQA train)} \\
% & & \multicolumn{1}{c}{GAP-VQA train)} \\
\midrule
LLaVA Bench & \textbf{87.9} & 82.2 \\
OCRBench & 72.4 & \textbf{73.1} \\
MME & 2025.61 & \textbf{2040.54} \\
RealWorldQA & \textbf{0.634} & 0.609 \\
MM-Vet & 51.422 & \textbf{51.789} \\
MMBench & \textbf{0.752} & 0.7422 \\
HallusionBench & 59.93 & \textbf{60.25} \\
TextVQA & 76.63 & \textbf{76.966} \\
MMMU val & 0.474 & \textbf{0.486} \\
DocVQA & \textbf{84.47} & 84.33 \\
\bottomrule
\end{tabular}
\caption{Performance on benchmarks of MiniCPM-Llama3-V-2.5-8B before and after fine-tuning on the AP-VQA train dataset.}
\label{tab:performance_comparison_updated}
\end{table}

\paragraph{Cross Model Evaluation}

Our cross-model evaluation reveals a compelling phenomenon \ref{tab:cross_model_perf1} \ref{tab:cross_model_perf2}: addressing knowledge gaps in one model architecture leads to improvements across different model sizes and architectures. 
This suggests a previously under-explored commonality in multimodal model deficiencies.
We observe that GAP-VQA training, initially designed for MiniCPM, significantly enhances both Qwen2-VL-7B and Qwen2-VL-2B models across various benchmarks. 
% Notably, improvements in visual reasoning tasks, in particular LLaVA Bench and MME, are consistent across model sizes, 
% indicating that our targeted knowledge infusion addresses shared limitations.

The effectiveness of GAP-VQA across different model architectures suggests fundamental, shared gaps in multimodal understanding. For instance, both 7B and 2B models show marked improvements in LLaVA Bench and MME, with the smaller 2B model exhibiting substantial relative gains. 
% This consistency across scales underscores the robustness of our approach.
Furthermore, we note that while enhancing performance on targeted tasks, our method preserves or improves capabilities on specialized benchmarks like OCRBench and DocVQA. 
This indicates that GAP-VQA complements existing model strengths while addressing common weaknesses.

% This observed phenomenon of shared weaknesses across models opens new avenues for efficient multimodal AI improvement. It suggests that targeted dataset curation can yield broad benefits across diverse model architectures, potentially streamlining the development of more capable and generalizable AI systems. Our findings highlight the potential for more unified approaches to multimodal model enhancement, where improvements designed for one architecture can have far-reaching impacts across the field.

\begin{table}[tbp]
\centering
\small
\begin{tabular}{lrr}
\toprule
Dataset & Qwen2-VL-7B (orig) & Qwen2-VL-7B (finetuned on GAP-VQA train) \\
\midrule
LLaVA Bench & 76.7 & \textbf{83.6} \\
OCRBench & 86.1 & \textbf{86.7} \\
MME & 2318.98 & \textbf{2332.71} \\
RealWorldQA & \textbf{0.699} & 0.690 \\
MM-Vet & 62.889 & \textbf{64.954} \\
MMBench & 0.808 & \textbf{0.815} \\
HallusionBench & 68.769 & 68.769 \\
TextVQA & \textbf{84.428} & 84.084 \\
MMMU val& 0.524 & \textbf{0.527} \\
DocVQA & 93.866 & \textbf{94.038} \\
\bottomrule
\end{tabular}
\caption{Performance Qwen2-VL-7B before and after fine-tuning on GAP-VQA train dataset.}
\label{tab:cross_model_perf1}
\end{table}

\begin{table}[htbp]
\centering
\small
\begin{tabular}{lrr}
\toprule
Dataset & Qwen2-VL-2B (orig) & Qwen2-VL-2B (finetuned on GAP-VQA train) \\
\midrule
LLaVA Bench & 52.6 & \textbf{57.9} \\
OCRBench & 81.2 & \textbf{81.4} \\
MME & 1881.92 & \textbf{1962.75} \\
RealWorldQA & \textbf{0.626} & 0.6156 \\
MM-Vet & 51.146 & \textbf{52.43} \\
MMBench & 0.729 & \textbf{0.732} \\
HallusionBench & 61.619 & \textbf{62.99} \\
TextVQA & 79.824 & \textbf{80.074} \\
MMMU val & 0.414 & \textbf{0.448} \\
DocVQA & 89.26 & \textbf{89.36} \\
\bottomrule
\end{tabular}
\caption{Performance of Qwen2-VL-2B before and after fine-tuning on GAP-VQA train dataset.}
\label{tab:cross_model_perf2}
\end{table}

% \section{Conclusion}
% The Gamified Adversarial Prompting (GAP) framework demonstrates significant potential for improving Visual Question Answering (VQA) models through targeted, human-in-the-loop data collection. Key findings include substantial performance gains for the MiniCPM-Llama3-V-2.5-8B model, cross-dataset generalization improvements, and cross-model benefits extending to Qwen2-VL models. The framework successfully engaged over 50,000 participants, with a significant portion showing high engagement and accuracy. However, limitations such as potential bias in data collection, the need for robust automated contribution detection, and questions about long-term participant engagement suggest areas for improvement. Future work should focus on refining gamification elements, expanding to diverse participant pools, and exploring applications beyond VQA. Despite these challenges, GAP presents a promising approach for continuous, scalable improvement of VQA models, leveraging human creativity and engagement in a way that could significantly advance the field of artificial intelligence.

\section{Discussion}
The Gamified Adversarial Prompting (GAP) framework represents a significant advance in the field of AI model improvement.
%for Visual Question Answering. 
Our results demonstrate substantial improvements in VQA across multiple benchmarks and models.
The implications of GAP extend beyond immediate performance gains, 
suggesting a new paradigm for continuous, scalable AI improvement that leverages human creativity and engagement. 
% As AI systems become increasingly integrated into various aspects of society, frameworks like GAP that enable rapid, targeted improvements will be crucial in ensuring these systems remain robust, reliable, and aligned with human needs.

GAP offers advantages over methods that rely on AI self-assessment or using
one AI model to evaluate another. 
Such approaches can be efficient, but they risk perpetuating
or even amplifying existing biases and errors. 
% When one AI model is used to assess or ”teach” another, there’s a potential for the evaluated model to simply reproduce the characteristics—including
% flaws—of the assessing model. This process can lead to a form of algorithmic echo chamber, where
% mistakes and biases are reinforced rather than corrected.
Furthermore, in cases where AI models are trained using outputs from other models without proper
attribution or permission, ethical concerns arise regarding the appropriation of intellectual property.
% It’s important to note that many leading AI models, including Claude and ChatGPT, have explicit
% licensing terms that prohibit the use of their outputs for training other AI models (for Commercial
% Purposes)
Our human-in-the-loop approach sidesteps these issues by leveraging human cognition and diverse
perspectives to drive model improvement. 
% This not only ensures a more robust and unbiased evaluation but also aligns with ethical standards in AI development and training. 
By relying on human
evaluation within a gamified framework, we respect legal and ethical boundaries while providing
a more transparent method for model improvement. 
This ensures that our model’s growth is built
on a foundation of original, human-verified data, rather than potentially restricted or problematic
AI-generated content.\\
Our future work will focus on enhancing the GAP framework through three key developments: (1) a visually fine-tuned language model for generating questions that the base LLM answers incorrectly, systematically identifying the blind spots; (2) a sophisticated probabilistic model incorporating player skill, image difficulty, response time, and fatigue to more accurately estimate LMM capabilities while controlling for confounding variables \ref{appendix:player_interaction_model}; and (3) exploration of GAP's applicability beyond LMMs to other AI domains, addressing domain-specific challenges and ethical considerations. These advancements aim to create a scalable, gamified approach for continuous improvement of AI systems through targeted human feedback.

\newpage

\bibliography{iclr2025_conference}

\begin{thebibliography}{64}
\providecommand{\natexlab}[1]{#1}
\providecommand{\url}[1]{\texttt{#1}}
\expandafter\ifx\csname urlstyle\endcsname\relax
  \providecommand{\doi}[1]{doi: #1}\else
  \providecommand{\doi}{doi: \begingroup \urlstyle{rm}\Url}\fi

\bibitem[Abdin et~al.(2024)Abdin, Jacobs, Awan, Aneja, Awadallah, Awadalla, Bach, Bahree, Bakhtiari, Behl, et~al.]{abdin2024phi}
Marah Abdin, Sam~Ade Jacobs, Ammar~Ahmad Awan, Jyoti Aneja, Ahmed Awadallah, Hany Awadalla, Nguyen Bach, Amit Bahree, Arash Bakhtiari, Harkirat Behl, et~al.
\newblock Phi-3 technical report: A highly capable language model locally on your phone.
\newblock \emph{arXiv preprint arXiv:2404.14219}, 2024.

\bibitem[Ahn et~al.(2008)Ahn, Maurer, McMillen, Abraham, and Blum]{ahn2008}
Luis Ahn, Benjamin Maurer, Colin McMillen, David Abraham, and Manuel Blum.
\newblock recaptcha: Human-based character recognition via web security measures.
\newblock \emph{Science (New York, N.Y.)}, 321:\penalty0 1465--8, 09 2008.
\newblock \doi{10.1126/science.1160379}.

\bibitem[Alayrac et~al.(2022)Alayrac, Donahue, Luc, Miech, Barr, Hasson, Lenc, Mensch, Millican, Reynolds, et~al.]{alayrac2022flamingo}
Jean-Baptiste Alayrac, Jeff Donahue, Pauline Luc, Antoine Miech, Iain Barr, Yana Hasson, Karel Lenc, Arthur Mensch, Katherine Millican, Malcolm Reynolds, et~al.
\newblock Flamingo: a visual language model for few-shot learning.
\newblock \emph{Advances in Neural Information Processing Systems}, 35:\penalty0 23716--23736, 2022.

\bibitem[Bai et~al.(2023)Bai, Bai, Yang, Wang, Tan, Wang, Lin, Zhou, and Zhou]{bai2023qwen}
Jinze Bai, Shuai Bai, Shusheng Yang, Shijie Wang, Sinan Tan, Peng Wang, Junyang Lin, Chang Zhou, and Jingren Zhou.
\newblock Qwen-vl: A versatile vision-language model for understanding, localization, text reading, and beyond.
\newblock \emph{arXiv preprint arXiv:2308.12966}, 2023.

\bibitem[Bartolo et~al.(2020)Bartolo, Roberts, Welbl, Riedel, and Stenetorp]{bartolo2020beat}
Max Bartolo, Alastair Roberts, Johannes Welbl, Sebastian Riedel, and Pontus Stenetorp.
\newblock Beat the ai: Investigating adversarial human annotation for reading comprehension.
\newblock \emph{Transactions of the Association for Computational Linguistics}, 8:\penalty0 662--678, 2020.

\bibitem[Brown et~al.(2020)Brown, Mann, Ryder, Subbiah, Kaplan, Dhariwal, Neelakantan, Shyam, Sastry, Askell, et~al.]{brown2020language}
Tom Brown, Benjamin Mann, Nick Ryder, Melanie Subbiah, Jared~D Kaplan, Prafulla Dhariwal, Arvind Neelakantan, Pranav Shyam, Girish Sastry, Amanda Askell, et~al.
\newblock Language models are few-shot learners.
\newblock \emph{Advances in neural information processing systems}, 33:\penalty0 1877--1901, 2020.

\bibitem[Cheah et~al.(2022)Cheah, Shimul, and Phau]{cheah2022motivations}
Isaac Cheah, Anwar~Sadat Shimul, and Ian Phau.
\newblock Motivations of playing digital games: A review and research agenda.
\newblock \emph{Psychology \& Marketing}, 39\penalty0 (5):\penalty0 937--950, 2022.

\bibitem[Chen et~al.(2023)Chen, Li, Dong, Zhang, He, Wang, Zhao, and Lin]{chen2023sharegpt4v}
Lin Chen, Jinsong Li, Xiaoyi Dong, Pan Zhang, Conghui He, Jiaqi Wang, Feng Zhao, and Dahua Lin.
\newblock Sharegpt4v: Improving large multi-modal models with better captions.
\newblock \emph{arXiv preprint arXiv:2311.12793}, 2023.

\bibitem[Dai et~al.(2023)Dai, Liu, Luo, Zhou, Dong, Zhang, Bai, Jiang, Tan, Bao, et~al.]{dai2023instructblip}
Wenliang Dai, Junnan Liu, Peng Luo, Huaishao Zhou, Yixiao Dong, Lu~Zhang, Yufeng Bai, Shuicheng Jiang, Descai Tan, Yue Bao, et~al.
\newblock Instructblip: Towards general-purpose vision-language models with instruction tuning.
\newblock \emph{arXiv preprint arXiv:2305.06500}, 2023.

\bibitem[Deci \& Ryan(2012)Deci and Ryan]{deci2012self}
Edward~L Deci and Richard~M Ryan.
\newblock Self-determination theory.
\newblock \emph{Handbook of theories of social psychology}, 1\penalty0 (20):\penalty0 416--436, 2012.

\bibitem[Duan et~al.(2024)Duan, Yang, Qiao, Fang, Chen, Liu, Dong, Zang, Zhang, Wang, Lin, and Chen]{duan2024vlmevalkit}
Haodong Duan, Junming Yang, Yuxuan Qiao, Xinyu Fang, Lin Chen, Yuan Liu, Xiaoyi Dong, Yuhang Zang, Pan Zhang, Jiaqi Wang, Dahua Lin, and Kai Chen.
\newblock Vlmevalkit: An open-source toolkit for evaluating large multi-modality models, 2024.
\newblock URL \url{https://arxiv.org/abs/2407.11691}.

\bibitem[Dubey et~al.(2024)Dubey, Jauhri, Pandey, Kadian, Al-Dahle, Letman, Mathur, Schelten, Yang, Fan, et~al.]{dubey2024llama}
Abhimanyu Dubey, Abhinav Jauhri, Abhinav Pandey, Abhishek Kadian, Ahmad Al-Dahle, Aiesha Letman, Akhil Mathur, Alan Schelten, Amy Yang, Angela Fan, et~al.
\newblock The llama 3 herd of models.
\newblock \emph{arXiv preprint arXiv:2407.21783}, 2024.

\bibitem[Eisenschlos et~al.(2021)Eisenschlos, Dhingra, Bulian, B{\"o}rschinger, and Boyd-Graber]{eisenschlos2021fool}
Julian Eisenschlos, Bhuwan Dhingra, Jannis Bulian, Benjamin B{\"o}rschinger, and Jordan Boyd-Graber.
\newblock Fool me twice: Entailment from wikipedia gamification.
\newblock \emph{arXiv preprint arXiv:2104.04725}, 2021.

\bibitem[Fisch et~al.(2019)Fisch, Talmor, Jia, Seo, Choi, and Chen]{fisch2019mrqa}
Adam Fisch, Alon Talmor, Robin Jia, Minjoon Seo, Eunsol Choi, and Danqi Chen.
\newblock Mrqa 2019 shared task: Evaluating generalization in reading comprehension.
\newblock In \emph{Proceedings of the 2nd Workshop on Machine Reading for Question Answering}, pp.\  1--13, 2019.

\bibitem[Fu et~al.(2023)Fu, Chen, Shen, Qin, Zhang, Lin, Qiu, Lin, Yang, Zheng, et~al.]{fu2023mme}
Chaoyou Fu, Peixian Chen, Yunhang Shen, Yulei Qin, Mengdan Zhang, Xu~Lin, Zhenyu Qiu, Wei Lin, Jinrui Yang, Xiawu Zheng, et~al.
\newblock Mme: A comprehensive evaluation benchmark for multimodal large language models.
\newblock \emph{arXiv preprint arXiv:2306.13394}, 2023.

\bibitem[Gao et~al.(2024)Gao, Zhang, Liu, Qiu, Huang, Lin, Zhao, Geng, Lin, Jin, et~al.]{gao2024sphinx}
Peng Gao, Renrui Zhang, Chris Liu, Longtian Qiu, Siyuan Huang, Weifeng Lin, Shitian Zhao, Shijie Geng, Ziyi Lin, Peng Jin, et~al.
\newblock Sphinx-x: Scaling data and parameters for a family of multi-modal large language models.
\newblock \emph{arXiv preprint arXiv:2402.05935}, 2024.

\bibitem[Gardner et~al.(2020)Gardner, Artzi, Basmov, Berant, Bogin, Chen, Dasigi, Dua, Elazar, Gottumukkala, et~al.]{gardner2020evaluating}
Matt Gardner, Yoav Artzi, Victoria Basmov, Jonathan Berant, Ben Bogin, Sihao Chen, Pradeep Dasigi, Dheeru Dua, Yanai Elazar, Ananth Gottumukkala, et~al.
\newblock Evaluating models' local decision boundaries via contrast sets.
\newblock In \emph{Findings of the Association for Computational Linguistics: EMNLP 2020}, pp.\  1307--1323, 2020.

\bibitem[Geva et~al.(2021)Geva, Khashabi, Segal, Khot, Roth, and Berant]{geva2021did}
Mor Geva, Daniel Khashabi, Elad Segal, Tushar Khot, Dan Roth, and Jonathan Berant.
\newblock Did aristotle use a laptop? a question answering benchmark with implicit reasoning strategies.
\newblock \emph{Transactions of the Association for Computational Linguistics}, 9:\penalty0 346--361, 2021.

\bibitem[Girdhar \& Ramanan(2019)Girdhar and Ramanan]{girdhar2019cater}
Rohit Girdhar and Deva Ramanan.
\newblock Cater: A diagnostic dataset for compositional actions and temporal reasoning.
\newblock \emph{arXiv preprint arXiv:1910.04744}, 2019.

\bibitem[Guan et~al.(2023)Guan, Liu, Wu, Xian, Li, Liu, Wang, Chen, Huang, Yacoob, Manocha, and Zhou]{guan2023hallusionbench}
Tianrui Guan, Fuxiao Liu, Xiyang Wu, Ruiqi Xian, Zongxia Li, Xiaoyu Liu, Xijun Wang, Lichang Chen, Furong Huang, Yaser Yacoob, Dinesh Manocha, and Tianyi Zhou.
\newblock Hallusionbench: An advanced diagnostic suite for entangled language hallucination \& visual illusion in large vision-language models, 2023.

\bibitem[Hong et~al.(2024)Hong, Wang, Ding, Yu, Lv, Wang, Cheng, Huang, Ji, Xue, Zhao, Yang, Gu, Zhang, Feng, Yin, Wang, Qi, Song, Zhang, Liu, Xu, Li, Dong, and Tang]{hong2024cogvlm2}
Wenyi Hong, Weihan Wang, Ming Ding, Wenmeng Yu, Qingsong Lv, Yan Wang, Yean Cheng, Shiyu Huang, Junhui Ji, Zhao Xue, Lei Zhao, Zhuoyi Yang, Xiaotao Gu, Xiaohan Zhang, Guanyu Feng, Da~Yin, Zihan Wang, Ji~Qi, Xixuan Song, Peng Zhang, Debing Liu, Bin Xu, Juanzi Li, Yuxiao Dong, and Jie Tang.
\newblock Cogvlm2: Visual language models for image and video understanding.
\newblock \emph{arXiv preprint arXiv:2408.16500}, 2024.

\bibitem[Hu et~al.(2022)Hu, Shen, Wallis, Allen-Zhu, Li, Wang, Wang, and Chen]{hu2022lora}
Edward~J Hu, Yelong Shen, Phillip Wallis, Zeyuan Allen-Zhu, Yuanzhi Li, Shean Wang, Lu~Wang, and Weizhu Chen.
\newblock Lo{RA}: Low-rank adaptation of large language models.
\newblock In \emph{International Conference on Learning Representations}, 2022.
\newblock URL \url{https://openreview.net/forum?id=nZeVKeeFYf9}.

\bibitem[Jia \& Liang(2017)Jia and Liang]{jia2017adversarial}
Robin Jia and Percy Liang.
\newblock Adversarial examples for evaluating reading comprehension systems.
\newblock In \emph{Proceedings of the 2017 Conference on Empirical Methods in Natural Language Processing}, pp.\  2021--2031, 2017.

\bibitem[Kanellopoulou et~al.(2020)Kanellopoulou, Giannakoulopoulos, et~al.]{kanellopoulou2020engage}
Catherine Kanellopoulou, Andreas Giannakoulopoulos, et~al.
\newblock Engage and conquer: An online empirical approach into whether intrinsic or extrinsic motivation leads to more enhanced students’ engagement.
\newblock \emph{Creative Education}, 11\penalty0 (02):\penalty0 143, 2020.

\bibitem[Kaushik et~al.(2020)Kaushik, Hovy, and Lipton]{kaushik2020learning}
Divyansh Kaushik, Eduard Hovy, and Zachary~C Lipton.
\newblock Learning the difference that makes a difference with counterfactually-augmented data.
\newblock In \emph{International Conference on Learning Representations}, 2020.

\bibitem[Kiela et~al.(2021)Kiela, Bartolo, Nie, Kaushik, Geiger, Wu, Vidgen, Prasad, Singh, Ringshia, et~al.]{kiela2021dynabench}
Douwe Kiela, Max Bartolo, Yixin Nie, Divyansh Kaushik, Atticus Geiger, Zhengxuan Wu, Bertie Vidgen, Grusha Prasad, Amanpreet Singh, Pratik Ringshia, et~al.
\newblock Dynabench: Rethinking benchmarking in nlp.
\newblock \emph{arXiv preprint arXiv:2104.14337}, 2021.

\bibitem[Kim et~al.(2024)Kim, Mitra, Chen, Rahman, and Zhang]{kim2024megannohumanllmcollaborativeannotation}
Hannah Kim, Kushan Mitra, Rafael~Li Chen, Sajjadur Rahman, and Dan Zhang.
\newblock Meganno+: A human-llm collaborative annotation system, 2024.
\newblock URL \url{https://arxiv.org/abs/2402.18050}.

\bibitem[LAION(2023)]{laion2023gpt4v}
LAION.
\newblock Laion/gpt4v-dataset.
\newblock \url{https://huggingface.co/datasets/laion/gpt4v-dataset}, 2023.

\bibitem[Lauren{\c{c}}on et~al.(2024)Lauren{\c{c}}on, Tronchon, Cord, and Sanh]{laurencon2024matters}
Hugo Lauren{\c{c}}on, Leo Tronchon, Matthieu Cord, and Victor Sanh.
\newblock What matters when building vision-language models?
\newblock \emph{arXiv preprint arXiv:2405.02246}, 2024.

\bibitem[Le~Bras et~al.(2020)Le~Bras, Swayamdipta, Bhagavatula, Zellers, Peters, Sabharwal, and Choi]{le2020adversarial}
Ronan Le~Bras, Swabha Swayamdipta, Chandra Bhagavatula, Rowan Zellers, Matthew~E Peters, Ashish Sabharwal, and Yejin Choi.
\newblock Adversarial filters of dataset biases.
\newblock In \emph{International Conference on Machine Learning}, pp.\  1078--1088. PMLR, 2020.

\bibitem[Li et~al.(2024{\natexlab{a}})Li, Zhang, Zhang, Guo, Zhang, Li, Zhang, Liu, and Li]{li2024llava}
Bo~Li, Kaichen Zhang, Hao Zhang, Dong Guo, Renrui Zhang, Feng Li, Yuanhan Zhang, Ziwei Liu, and Chunyuan Li.
\newblock Llava-next: Improved reasoning, ocr, and world knowledge.
\newblock \emph{arXiv preprint arXiv:2401.13236}, 2024{\natexlab{a}}.

\bibitem[Li et~al.(2023{\natexlab{a}})Li, Wong, Zhang, Usuyama, Liu, Yang, Naumann, Poon, and Gao]{li2023llava}
Chunyuan Li, Cliff Wong, Sheng Zhang, Naoto Usuyama, Haotian Liu, Jianwei Yang, Tristan Naumann, Hoifung Poon, and Jianfeng Gao.
\newblock Llava-med: Training a large language-and-vision assistant for biomedicine in one day.
\newblock \emph{arXiv preprint arXiv:2306.00890}, 2023{\natexlab{a}}.

\bibitem[Li et~al.(2024{\natexlab{b}})Li, Wang, Zhu, Kuo, Xu, Chen, Jain, Shi, and Wen]{li2024cumo}
Jiachen Li, Xinyao Wang, Sijie Zhu, Chia-Wen Kuo, Lu~Xu, Fan Chen, Jitesh Jain, Humphrey Shi, and Longyin Wen.
\newblock Cumo: Scaling multimodal llm with co-upcycled mixture-of-experts.
\newblock \emph{arXiv preprint arXiv:2405.05949}, 2024{\natexlab{b}}.

\bibitem[Li et~al.(2023{\natexlab{b}})Li, Zhang, Li, Chen, Chen, Cheng, Wang, Zhou, and Xiao]{li2023quantity}
Ming Li, Yong Zhang, Zhitao Li, Jiuhai Chen, Lichang Chen, Ning Cheng, Jianzong Wang, Tianyi Zhou, and Jing Xiao.
\newblock From quantity to quality: Boosting llm performance with self-guided data selection for instruction tuning.
\newblock \emph{arXiv preprint arXiv:2308.12032}, 2023{\natexlab{b}}.

\bibitem[Li et~al.(2023{\natexlab{c}})Li, Du, Zhou, Wang, Zhao, and Wen]{li2023evaluating}
Yifan Li, Yifan Du, Kun Zhou, Jinpeng Wang, Wayne~Xin Zhao, and Ji-Rong Wen.
\newblock Evaluating object hallucination in large vision-language models.
\newblock \emph{arXiv preprint arXiv:2305.10355}, 2023{\natexlab{c}}.

\bibitem[Lin et~al.(2014)Lin, Maire, Belongie, Hays, Perona, Ramanan, Doll{\'a}r, and Zitnick]{lin2014microsoft}
Tsung-Yi Lin, Michael Maire, Serge Belongie, James Hays, Pietro Perona, Deva Ramanan, Piotr Doll{\'a}r, and C~Lawrence Zitnick.
\newblock Microsoft coco: Common objects in context.
\newblock In \emph{Computer Vision--ECCV 2014: 13th European Conference, Zurich, Switzerland, September 6-12, 2014, Proceedings, Part V 13}, pp.\  740--755. Springer, 2014.

\bibitem[Liu et~al.(2023{\natexlab{a}})Liu, Li, Wu, and Lee]{liu2023visual}
Haotian Liu, Chunyuan Li, Qingyang Wu, and Yong~Jae Lee.
\newblock Visual instruction tuning.
\newblock \emph{arXiv preprint arXiv:2304.08485}, 2023{\natexlab{a}}.

\bibitem[Liu et~al.(2023{\natexlab{b}})Liu, Duan, Zhang, Li, Zhang, Zhao, Yuan, Wang, He, Liu, et~al.]{liu2023mmbench}
Yuan Liu, Haodong Duan, Yuanhan Zhang, Bo~Li, Songyang Zhang, Wangbo Zhao, Yike Yuan, Jiaqi Wang, Conghui He, Ziwei Liu, et~al.
\newblock Mmbench: Is your multi-modal model an all-around player?
\newblock \emph{arXiv preprint arXiv:2307.06281}, 2023{\natexlab{b}}.

\bibitem[Liu et~al.(2024)Liu, Li, Huang, Yang, Yu, Li, Yin, lin Liu, Jin, and Bai]{liu2024hiddenmysteryocrlarge}
Yuliang Liu, Zhang Li, Mingxin Huang, Biao Yang, Wenwen Yu, Chunyuan Li, Xucheng Yin, Cheng lin Liu, Lianwen Jin, and Xiang Bai.
\newblock On the hidden mystery of ocr in large multimodal models, 2024.
\newblock URL \url{https://arxiv.org/abs/2305.07895}.

\bibitem[Lu et~al.(2024)Lu, Liu, Zhang, Wang, Dong, Liu, Sun, Ren, Li, Sun, et~al.]{lu2024deepseek}
Haoyu Lu, Wen Liu, Bo~Zhang, Bingxuan Wang, Kai Dong, Bo~Liu, Jingxiang Sun, Tongzheng Ren, Zhuoshu Li, Yaofeng Sun, et~al.
\newblock Deepseek-vl: Towards real-world vision-language understanding.
\newblock \emph{arXiv preprint arXiv:2403.05525}, 2024.

\bibitem[Mahdisoltani et~al.(2018)Mahdisoltani, Berger, Gharbieh, Fleet, and Memisevic]{mahdisoltani2018effectivenesstaskgranularitytransfer}
Farzaneh Mahdisoltani, Guillaume Berger, Waseem Gharbieh, David Fleet, and Roland Memisevic.
\newblock On the effectiveness of task granularity for transfer learning, 2018.
\newblock URL \url{https://arxiv.org/abs/1804.09235}.

\bibitem[Materzynska et~al.(2020)Materzynska, Xiao, Herzig, Xu, Wang, and Darrell]{CVPR2020_SomethingElse}
Joanna Materzynska, Tete Xiao, Roei Herzig, Huijuan Xu, Xiaolong Wang, and Trevor Darrell.
\newblock Something-else: Compositional action recognition with spatial-temporal interaction networks.
\newblock In \emph{CVPR}, 2020.

\bibitem[Mathew et~al.(2021)Mathew, Karatzas, and Jawahar]{mathew2021docvqa}
Minesh Mathew, Dimosthenis Karatzas, and CV~Jawahar.
\newblock Docvqa: A dataset for vqa on document images.
\newblock In \emph{Proceedings of the IEEE/CVF winter conference on applications of computer vision}, pp.\  2200--2209, 2021.

\bibitem[Raina et~al.(2024)Raina, Liusie, and Gales]{raina2024llm}
Vyas Raina, Adian Liusie, and Mark Gales.
\newblock Is llm-as-a-judge robust? investigating universal adversarial attacks on zero-shot llm assessment.
\newblock \emph{arXiv preprint arXiv:2402.14016}, 2024.

\bibitem[Rajpurkar et~al.(2018)Rajpurkar, Jia, and Liang]{rajpurkar2018know}
Pranav Rajpurkar, Robin Jia, and Percy Liang.
\newblock Know what you don't know: Unanswerable questions for squad.
\newblock In \emph{Proceedings of the 56th Annual Meeting of the Association for Computational Linguistics (Volume 2: Short Papers)}, pp.\  784--789, 2018.

\bibitem[Ribeiro et~al.(2018)Ribeiro, Singh, and Guestrin]{ribeiro2018semantically}
Marco~Tulio Ribeiro, Sameer Singh, and Carlos Guestrin.
\newblock Semantically equivalent adversarial rules for debugging nlp models.
\newblock In \emph{Proceedings of the 56th Annual Meeting of the Association for Computational Linguistics (Volume 1: Long Papers)}, pp.\  856--865, 2018.

\bibitem[Sakaguchi et~al.(2020)Sakaguchi, Le~Bras, Bhagavatula, and Choi]{sakaguchi2020winogrande}
Keisuke Sakaguchi, Ronan Le~Bras, Chandra Bhagavatula, and Yejin Choi.
\newblock Winogrande: An adversarial winograd schema challenge at scale.
\newblock In \emph{Proceedings of the AAAI Conference on Artificial Intelligence}, pp.\  8732--8740, 2020.

\bibitem[Sap et~al.(2019)Sap, Rashkin, Chen, Le~Bras, and Choi]{sap2019social}
Maarten Sap, Hannah Rashkin, Derek Chen, Ronan Le~Bras, and Yejin Choi.
\newblock Social iqa: Commonsense reasoning about social interactions.
\newblock In \emph{Proceedings of the 2019 Conference on Empirical Methods in Natural Language Processing and the 9th International Joint Conference on Natural Language Processing (EMNLP-IJCNLP)}, pp.\  4463--4473, 2019.

\bibitem[Shnarch et~al.(2022)Shnarch, Halfon, Gera, Danilevsky, Katsis, Choshen, Cooper, Epelboim, Zhang, Wang, Yip, Ein-Dor, Dankin, Shnayderman, Aharonov, Li, Liberman, Slesarev, Newton, Ofek-Koifman, Slonim, and Katz]{shnarch2022labelsleuth}
Eyal Shnarch, Alon Halfon, Ariel Gera, Marina Danilevsky, Yannis Katsis, Leshem Choshen, Martin~Santillan Cooper, Dina Epelboim, Zheng Zhang, Dakuo Wang, Lucy Yip, Liat Ein-Dor, Lena Dankin, Ilya Shnayderman, Ranit Aharonov, Yunyao Li, Naftali Liberman, Philip~Levin Slesarev, Gwilym Newton, Shila Ofek-Koifman, Noam Slonim, and Yoav Katz.
\newblock {L}abel {S}leuth: From unlabeled text to a classifier in a few hours.
\newblock In \emph{Proceedings of the 2022 Conference on Empirical Methods in Natural Language Processing ({EMNLP}): System Demonstrations}, pp.\  159--168, Abu Dhabi, UAE, dec 2022. Association for Computational Linguistics.
\newblock URL \url{https://aclanthology.org/2022.emnlp-demos.16}.

\bibitem[Talmor \& Berant(2019)Talmor and Berant]{talmor2019multiqa}
Alon Talmor and Jonathan Berant.
\newblock Multiqa: An empirical investigation of generalization and transfer in reading comprehension.
\newblock In \emph{Proceedings of the 57th Annual Meeting of the Association for Computational Linguistics}, pp.\  4911--4921, 2019.

\bibitem[Talmor et~al.(2021)Talmor, Yoran, Le~Bras, Bhagavatula, Goldberg, Choi, and Berant]{talmor2021commonsenseqa}
Alon Talmor, Ori Yoran, Ronan Le~Bras, Chandra Bhagavatula, Yoav Goldberg, Yejin Choi, and Jonathan Berant.
\newblock Commonsenseqa 2.0: Exposing the limits of ai through gamification.
\newblock \emph{arXiv preprint arXiv:2201.05320}, 2021.

\bibitem[Tkachenko et~al.(2020-2024)Tkachenko, Malyuk, Holmanyuk, and Liubimov]{labelstudio}
Maxim Tkachenko, Mikhail Malyuk, Andrey Holmanyuk, and Nikolai Liubimov.
\newblock {Label Studio}: Data labeling software, 2020-2024.
\newblock URL \url{https://github.com/HumanSignal/label-studio}.
\newblock Open source software available from https://github.com/HumanSignal/label-studio.

\bibitem[Wang et~al.(2024)Wang, Bai, Tan, Wang, Fan, Bai, Chen, Liu, Wang, Ge, et~al.]{wang2024qwen2}
Peng Wang, Shuai Bai, Sinan Tan, Shijie Wang, Zhihao Fan, Jinze Bai, Keqin Chen, Xuejing Liu, Jialin Wang, Wenbin Ge, et~al.
\newblock Qwen2-vl: Enhancing vision-language model's perception of the world at any resolution.
\newblock \emph{arXiv preprint arXiv:2409.12191}, 2024.

\bibitem[Yao et~al.(2024)Yao, Yu, Zhang, Wang, Cui, Zhu, Cai, Li, Zhao, He, et~al.]{yao2024minicpmv}
Yuan Yao, Tianyu Yu, Ao~Zhang, Chongyi Wang, Junbo Cui, Hongji Zhu, Tianchi Cai, Haoyu Li, Weilin Zhao, Zhihui He, et~al.
\newblock Minicpm-v: A gpt-4v level mllm on your phone.
\newblock \emph{arXiv preprint arXiv:2408.01800}, 2024.

\bibitem[Yu et~al.(2023)Yu, Yang, Li, Wang, Lin, Liu, Wang, and Wang]{yu2023mm}
Weihao Yu, Zhengyuan Yang, Linjie Li, Jianfeng Wang, Kevin Lin, Zicheng Liu, Xinchao Wang, and Lijuan Wang.
\newblock Mm-vet: Evaluating large multimodal models for integrated capabilities.
\newblock \emph{arXiv preprint arXiv:2308.02490}, 2023.

\bibitem[Yu et~al.(2024)Yu, Yang, Li, Wang, Lin, Liu, Wang, and Wang]{yu2024mm}
Weihao Yu, Zhengyuan Yang, Linjie Li, Jianfeng Wang, Kevin Lin, Zicheng Liu, Xinchao Wang, and Lijuan Wang.
\newblock Mm-vet: Evaluating large multimodal models for integrated capabilities.
\newblock In \emph{International conference on machine learning}. PMLR, 2024.

\bibitem[Yuan et~al.(2023)Yuan, Chen, Cui, Gao, Zou, Cheng, Ji, Liu, and Sun]{yuan2023revisiting}
Lifan Yuan, Yangyi Chen, Ganqu Cui, Hongcheng Gao, Fangyuan Zou, Xingyi Cheng, Heng Ji, Zhiyuan Liu, and Maosong Sun.
\newblock Revisiting out-of-distribution robustness in nlp: Benchmarks, analysis, and llms evaluations.
\newblock \emph{Advances in Neural Information Processing Systems}, 36:\penalty0 58478--58507, 2023.

\bibitem[Yuan et~al.(2019)Yuan, He, Zhu, and Li]{yuan2019adversarial}
Xiaoyong Yuan, Pan He, Qile Zhu, and Xiaolin Li.
\newblock Adversarial examples: Attacks and defenses for deep learning.
\newblock \emph{IEEE transactions on neural networks and learning systems}, 30\penalty0 (9):\penalty0 2805--2824, 2019.

\bibitem[Yue et~al.(2024)Yue, Ni, Zhang, Zheng, Liu, Zhang, Stevens, Jiang, Ren, Sun, Wei, Yu, Yuan, Sun, Yin, Zheng, Yang, Liu, Huang, Sun, Su, and Chen]{yue2023mmmu}
Xiang Yue, Yuansheng Ni, Kai Zhang, Tianyu Zheng, Ruoqi Liu, Ge~Zhang, Samuel Stevens, Dongfu Jiang, Weiming Ren, Yuxuan Sun, Cong Wei, Botao Yu, Ruibin Yuan, Renliang Sun, Ming Yin, Boyuan Zheng, Zhenzhu Yang, Yibo Liu, Wenhao Huang, Huan Sun, Yu~Su, and Wenhu Chen.
\newblock Mmmu: A massive multi-discipline multimodal understanding and reasoning benchmark for expert agi.
\newblock In \emph{Proceedings of CVPR}, 2024.

\bibitem[Zellers et~al.(2018)Zellers, Bisk, Schwartz, and Choi]{zellers2018swag}
Rowan Zellers, Yonatan Bisk, Roy Schwartz, and Yejin Choi.
\newblock Swag: A large-scale adversarial dataset for grounded commonsense inference.
\newblock In \emph{Proceedings of the 2018 Conference on Empirical Methods in Natural Language Processing}, pp.\  93--104, 2018.

\bibitem[Zellers et~al.(2019)Zellers, Holtzman, Bisk, Farhadi, and Choi]{zellers2019hellaswag}
Rowan Zellers, Ari Holtzman, Yonatan Bisk, Ali Farhadi, and Yejin Choi.
\newblock Hellaswag: Can a machine really finish your sentence?
\newblock In \emph{Proceedings of the 57th Annual Meeting of the Association for Computational Linguistics}, pp.\  4791--4800, 2019.

\bibitem[Zhang et~al.(2023)Zhang, Dong, Wang, Cao, Xu, Ouyang, Zhao, Ding, Zhang, Duan, et~al.]{zhang2023internlm}
Pan Zhang, Xiaoyi Dong, Bin Wang, Yuhang Cao, Chao Xu, Linke Ouyang, Zhiyuan Zhao, Shuangrui Ding, Songyang Zhang, Haodong Duan, et~al.
\newblock Internlm-xcomposer: A vision-language large model for advanced text-image comprehension and composition.
\newblock \emph{arXiv preprint arXiv:2309.15112}, 2023.

\bibitem[Zheng et~al.(2024)Zheng, Zhang, Zhang, Ye, Luo, Feng, and Ma]{zheng2024llamafactory}
Yaowei Zheng, Richong Zhang, Junhao Zhang, Yanhan Ye, Zheyan Luo, Zhangchi Feng, and Yongqiang Ma.
\newblock Llamafactory: Unified efficient fine-tuning of 100+ language models.
\newblock In \emph{Proceedings of the 62nd Annual Meeting of the Association for Computational Linguistics (Volume 3: System Demonstrations)}, Bangkok, Thailand, 2024. Association for Computational Linguistics.
\newblock URL \url{http://arxiv.org/abs/2403.13372}.

\bibitem[Zhou et~al.(2024)Zhou, Liu, Xu, Iyer, Sun, Mao, Ma, Efrat, Yu, Yu, et~al.]{zhou2024lima}
Chunting Zhou, Pengfei Liu, Puxin Xu, Srinivasan Iyer, Jiao Sun, Yuning Mao, Xuezhe Ma, Avia Efrat, Ping Yu, Lili Yu, et~al.
\newblock Lima: Less is more for alignment.
\newblock \emph{Advances in Neural Information Processing Systems}, 36, 2024.

\end{thebibliography}
\bibliographystyle{iclr2025_conference}

\newpage

\appendix
\section{Appendix}
\subsection{Player Participation and Performance}
Figure~\ref{fig:user_participation} reveals key insights into user participation patterns. 69.68\% of accounts show no weekly participation, likely including airdrop farming bots---automated programs common in blockchain projects that perform minimal actions to qualify for potential rewards. Despite this, 30.32\% of users actively engage at least once per week, with 2.68\% participating in multiple weekly sessions.

Figure~\ref{fig:user_participation}(b) demonstrates that when users do participate, they overwhelmingly interact with all 10 images in a session, as evidenced by the pronounced spike at the 10-image mark. This suggests that the game design effectively encourages thorough engagement once a session begins.

The active user base, while smaller, provides valuable data on the game's appeal and effectiveness. The high completion rate of sessions and the presence of repeat participants indicate strong engagement among active users, showcasing the game's potential to retain and involve players consistently.

\begin{figure}[htbp]
    \centering
    \begin{minipage}{0.49\textwidth}
        \centering
        \begin{tikzpicture}[scale=1.5]
        \definecolor{myblue}{RGB}{0,114,178}
        \pgfmathsetmacro{\radius}{1.2}
        \newcommand{\pieSlice}[3]{
            \pgfmathsetmacro{\endangle}{#1*3.6+\lastangle}
            \pgfmathsetmacro{\midangle}{(\lastangle+\endangle)/2}
            \draw[fill=#2, draw=white, line width=0.5pt] (0,0) -- (\lastangle:\radius) arc (\lastangle:\endangle:\radius) -- cycle;
            \node[font=\tiny, text=white] at (\midangle:\radius*0.7) {#1\%};
            \pgfmathsetmacro{\lastangle}{\endangle}
        }
        \def\lastangle{0}
        \pieSlice{69.68}{myblue!30}{0}
        \pieSlice{27.75}{myblue!60}{1}
        \pieSlice{2.68}{myblue!90}{2+}

        % Legend
        \node[font=\footnotesize, align=left, anchor=north west] at (1.5, 1.2) {
            \textcolor{myblue!30}{$\blacksquare$} 0 sessions \\
            \textcolor{myblue!60}{$\blacksquare$} 1 session \\
            \textcolor{myblue!90}{$\blacksquare$} 2+ sessions
        };
        \end{tikzpicture}
        \caption*{(a)}
    \end{minipage}
    \hfill
    \begin{minipage}{0.49\textwidth}
        \centering
        % Adjust the bar width and axis width
        \begin{tikzpicture}
        \begin{axis}[
            width=\textwidth,
            height=5cm,
            ybar,
            bar width=0.4cm,
            xlabel={Num Images Interacted},
            xlabel style={font=\small},
            ylabel={Fraction of active sessions},
            ylabel style={font=\small},
            xticklabels={1,2,3,4,5,6,7,8,9,10},
            xtick=data,
            x tick label style={rotate=45,anchor=east,font=\footnotesize},
            ymajorgrids=true,
            grid style={dotted,color=gray!30},
            ylabel near ticks,
            ymin=0,
            ymax=0.7,
            scaled y ticks=false,
            yticklabel style={/pgf/number format/fixed, /pgf/number format/precision=2, font=\footnotesize},
            axis background/.style={fill=mygray!30},
            legend pos=north east,
            legend style={font=\footnotesize, fill=white, draw=none},
            tick label style={font=\footnotesize},
        ]
        \addplot[fill=myblue] coordinates {
            (1, 0.1356673961)
            (2, 0.0841235108)
            (3, 0.0333090201)
            (4, 0.0284463895)
            (5, 0.0150741551)
            (6, 0.0121565767)
            (7, 0.0063214198)
            (8, 0.0072939460)
            (9, 0.0070508144)
            (10, 0.6705567771)
        };
        \end{axis}
        \end{tikzpicture}
        \caption*{(b)}
    \end{minipage}%
    \caption{(a) Distribution of user participation in sessions per week. (b) Distribution of the number of images interacted per session.}
    \label{fig:user_participation}
\end{figure}

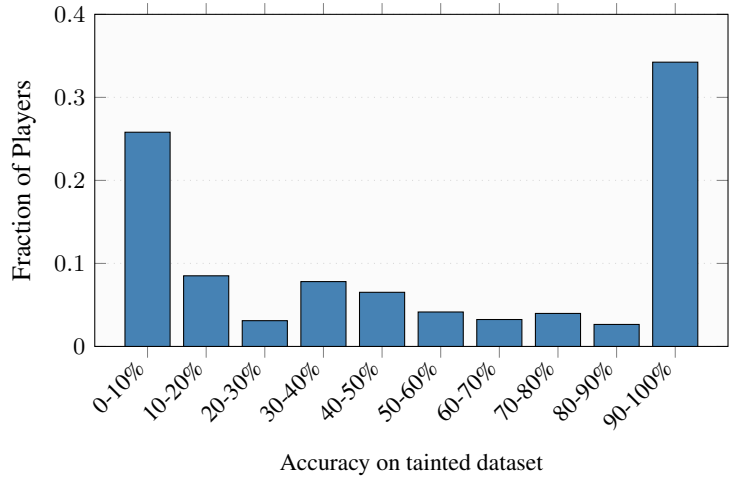
\begin{figure}[h]
\centering
\begin{tikzpicture}
\begin{axis}[
    width=10cm,
    height=6cm,
    ybar,
    bar width=0.6cm,
    xlabel={Accuracy on tainted dataset},
    xlabel style={font=\small},
    ylabel={Fraction of Players},
    ylabel style={font=\small},  % Changed to match xlabel
    xticklabels={0-10\%,10-20\%,20-30\%,30-40\%,40-50\%,50-60\%,60-70\%,70-80\%,80-90\%,90-100\%},
    xtick=data,
    x tick label style={rotate=45,anchor=east,font=\footnotesize},
    ymajorgrids=true,
    grid style={dotted,color=gray!30},
    ylabel near ticks,
    ymin=0,
    ymax=0.4,
    scaled y ticks=false,
    yticklabel style={/pgf/number format/fixed, /pgf/number format/precision=2, font=\footnotesize},
    axis background/.style={fill=mygray!30},
    legend pos=north west,
    legend style={font=\footnotesize, fill=white, draw=none},
    tick label style={font=\footnotesize},
]

\addplot[fill=myblue] coordinates {
    (1, 0.2580195258)
    (2, 0.0850767085)
    (3, 0.0310320781)
    (4, 0.0781032078)
    (5, 0.0652022315)
    (6, 0.0414923291)
    (7, 0.0324267782)
    (8, 0.0397489539)
    (9, 0.0264993026)
    (10, 0.3423988842)
};

\end{axis}
\end{tikzpicture}
\caption{Figure illustrates the bimodal distribution of $P (M = 0 | H = 0, D = \text{Tainted})$, revealing that participants tend to cluster at either high or low accuracy levels, as evaluated on the tainted dataset, with fewer achieving mid-range performance
}
\label{fig:accuracy}
\end{figure}

Figure \ref{fig:accuracy} presents a striking bimodal distribution of participant accuracy on the tainted dataset. A substantial 35\% of participants achieved exceptional accuracy in the 90-100\% range, demonstrating the task's solvability and the existence of highly effective strategies. Conversely, approximately 26\% scored in the 0-10\% range, indicating the task's ability to effectively differentiate skill levels. This U-shaped distribution, with peaks at both extremes and lower representation in middle ranges, suggests an optimal difficulty balance.

\subsection{GPT-4V Evaluation}
\label{appendix:gpt-4v_evaluation}
We used the following prompt to evaluate answers on GAP-VQA-val using GPT-4V:
\begin{tcolorbox}[
  colback=gray!10,
  colframe=gray!50,
  boxrule=0.5pt,
  arc=4pt,
  outer arc=4pt,
]
Please evaluate the model's answer based on the following criteria compared to the correct answer:
\begin{enumerate}
    \item Correct Answer: ``\{premise\}''
    \item Model Answer: ``\{hypothesis\}''
\end{enumerate}

\textbf{Criteria for evaluation:}
\begin{itemize}
    \item \textbf{Existence:} Does the model's answer correctly identify the existence or non-existence of objects or elements described in the correct answer?
    \item \textbf{Position:} Does the model's answer accurately describe the position or location of objects or elements as stated in the correct answer?
    \item \textbf{Count:} Does the model's answer correctly state the number of objects or elements mentioned in the correct answer?
    \item \textbf{Color:} Does the model's answer accurately describe the color of objects or elements as indicated in the correct answer?
\end{itemize}

Assign a score from 0 to 1 based on how well the model's answer meets these criteria:
\begin{itemize}
    \item A score of ``1'' means the model's answer fully meets all criteria, accurately reflecting existence, position, count, and color as described in the correct answer.
    \item A score of ``0'' means the model's answer fails to meet any of the criteria, showing no alignment with the correct answer.
    \item Scores between ``0'' and ``1'' should reflect partial correctness, where the model's answer meets some criteria but not all, or has minor inaccuracies.
\end{itemize}

Carefully consider each criterion before deciding. What is the appropriate score (between 0 and 1) that best represents the factual correctness of the model's answer? Just return the score as a single number.

\end{tcolorbox}

\begin{table}[t]
\centering
\small
\begin{tabular}{p{0.3\textwidth}p{0.7\textwidth}}
\toprule
\textbf{Category} & \textbf{Example Questions} \\
\midrule
Object Recognition & "What is the object on the table?" \\
                   & "Is there a cat in the picture?" \\
\midrule
Scene Understanding & "Where is this place?" \\
                    & "Is this an indoor or outdoor scene?" \\
\midrule
Activity Recognition & "What is the person doing?" \\
                     & "Are the people playing a sport?" \\
\midrule
Object Attributes & "What color is the car?" \\
                  & "Is the cup made of glass?" \\
\midrule
Spatial Relationships & "Is the apple next to the book?" \\
                      & "How many people are in front of the bus?" \\
\midrule
Counting & "How many dogs are in the park?" \\
         & "How many windows does the building have?" \\
\midrule
Text Recognition & "What does the sign say?" \\
                 & "Can you read the text on the billboard?" \\
\midrule
Emotion Recognition & "Is the person happy or sad?" \\
                    & "Does the dog look scared?" \\
\midrule
Weather and Environment & "Is it raining?" \\
                        & "Is the sky clear or cloudy?" \\
\midrule
Temporal Inference & "Is this image taken during the day or at night?" \\
                   & "Is it summer or winter in this picture?" \\
\midrule
Cultural or Social Context & "Is this a traditional wedding?" \\
                           & "What festival is being celebrated?" \\
\midrule
Art and Aesthetics & "What style of painting is this?" \\
                   & "Is this photograph black and white or colored?" \\
\midrule
Anomaly Detection & "Is there anything strange in this picture?" \\
                  & "Is there an error in this scene?" \\
\bottomrule
\end{tabular}
\caption{Categories of Visual Question Answering (VQA) Tasks}
\label{tab:vqa-categories}
\end{table}

\subsection{Player Interaction Model} \label{appendix:player_interaction_model}

The probabilistic model for evaluating user-generated questions designed to identify AI mistakes in image analysis
incorporates player ability, image difficulty, time pressure, and fatigue to estimate the probability of a player submitting a question that reveals an AI error.

\paragraph{Formal Definition}
We define our VQA model as a function $f: \mathcal{C} \times \mathcal{Q} \times \Theta \times \Omega \to \mathcal{R}$. Let's examine each space in detail:

\begin{itemize}
    \item $\mathcal{C}$ (Context Space): This represents all possible input images or visual contexts.
    \textit{Example:} $\mathcal{C}$ includes all possible digital images, such as photographs of landmarks, diagrams of scientific concepts, or snapshots of everyday scenes.

    \item $\mathcal{Q}$ (Question Space): This encompasses all possible questions that can be asked about the visual contexts.
    \textit{Example:} $\mathcal{Q}$ includes questions like "What color is the car?", "How many people are in the image?", or "What type of architecture is shown in this building?".

    \item $\Theta$ (Parameter Space): This represents all possible configurations of the model's parameters.
    \textit{Example:} For a neural network, $\Theta$ would include all possible weight and bias values for all layers of the network.

    \item $\Omega$ (Probability Space): This captures the stochastic nature of the model, representing all possible random outcomes.
    \textit{Example:} $\Omega$ could represent the randomness in dropout layers, or the variability in outputs due to temperature scaling in the final softmax layer.

    \item $\mathcal{R}$ (Response Space): This is the space of all possible model outputs or responses.
    \textit{Example:} $\mathcal{R}$ could include textual answers like "The car is red", numerical answers like "There are 3 people", or more complex responses like "The building exhibits Gothic architecture".
\end{itemize}

Additionally, we define:

\begin{itemize}
    \item $\mathcal{A}$ (Answer Space): A subset of $\mathcal{R}$ that contains all correct answers.
    \textit{Example:} For the question "What color is the sky?", $\mathcal{A}$ might include \{"Blue", "Azure", "Cyan"\} depending on the specific image.

    \item $g: \mathcal{C} \times \mathcal{Q} \to 2^\mathcal{A}$ (Correct Answer Function): This function maps a context-question pair to a set of correct answers.
    \textit{Example:} $g$(image of Eiffel Tower, "Where is this landmark located?") = \{"Paris", "France", "The capital of France"\}
\end{itemize}

We define \textbf{factual equivalence} $\sim$ as a relation on $\mathcal{R}$. For example:

\begin{itemize}
    \item "A dozen eggs" $\sim$ "12 eggs"
    \item "The capital of France" $\sim$ "Paris"
    \item "H2O" $\sim$ "Water"
\end{itemize}

An \textbf{adversarial sample} is a triplet $(c, q, \theta) \in \mathcal{C} \times \mathcal{Q} \times \Theta$ such that:

\begin{equation}
P(\forall r \in g(c,q), f(c, q; \theta, \omega) \not\sim r) > \epsilon
\end{equation}

where $\omega \in \Omega$ and $\epsilon$ is a threshold probability.

\textit{Example:} For a VQA model, an adversarial sample could be (image of the Eiffel Tower, "In which city is this monument located?", $\theta$) where the model consistently fails to answer "Paris" or its factual equivalents.

An \textbf{informative datapoint} is derived from an adversarial sample and consists of $(c, q, A)$ where $A = g(c,q)$. 

\textit{Example:} Following the previous example, the corresponding informative datapoint would be (image of the Eiffel Tower, "In which city is this monument located?", \{"Paris", "The capital of France"\}).

Let $L: \Theta \times \mathcal{C} \times \mathcal{Q} \times 2^\mathcal{A} \to \mathbb{R}$ be our loss function. The expected change in loss for a datapoint $(c, q, A)$ is:

\begin{equation}
\Delta L(\theta, c, q, A) = E_{\omega \in \Omega}[L(\theta_{new}, c, q, A) - L(\theta, c, q, A)]
\end{equation}

where $\theta_{new}$ is the updated parameter after training.

We define the global expected loss as:

\begin{equation}
\mathcal{L}_{global}(\theta) = E_{(c,q) \sim P(c,q)}[L(\theta, c, q, g(c,q))]
\end{equation}

where $P(c,q)$ is the true distribution of context-question pairs in the domain.

An \textbf{informative dataset} $D_I$ is a collection of informative datapoints:

\begin{equation}
D_I = \{(c_i, q_i, A_i) | (c_i, q_i) \in \mathcal{C} \times \mathcal{Q}, A_i = g(c_i, q_i), E(c_i, q_i, A_i) > \tau\}
\end{equation}

where $\tau$ is a threshold for datapoint effectiveness.

\textit{Example:} An informative dataset might include various challenging VQA pairs, such as:
\begin{itemize}
    \item (image of DNA double helix, "What molecule is this?", \{"DNA", "Deoxyribonucleic acid"\})
    \item (image of a quasar, "What astronomical object is this?", \{"Quasar", "Quasi-stellar object"\})
    \item (image of the Mona Lisa, "Who painted this masterpiece?", \{"Leonardo da Vinci", "da Vinci"\})
\end{itemize}

Let $D_U$ be an uninformative (random) dataset. The global improvement for each dataset is:

\begin{align}
\Delta \mathcal{L}_{global,I} &= \mathcal{L}_{global}(\theta) - \mathcal{L}_{global}(\theta_I) \\
\Delta \mathcal{L}_{global,U} &= \mathcal{L}_{global}(\theta) - \mathcal{L}_{global}(\theta_U)
\end{align}

We argue that $\Delta \mathcal{L}_{global,I} > \Delta \mathcal{L}_{global,U}$ for the following reasons:

\begin{enumerate}
    \item \textbf{Larger Gradient Magnitude:}
    \begin{equation}
    E[\|\nabla_\theta L(\theta, c, q, A)\| | (c,q,A) \in D_I] > E[\|\nabla_\theta L(\theta, c, q, A)\| | (c,q,A) \in D_U]
    \end{equation}

    \item \textbf{Higher Information Gain:}
    \begin{equation}
    E[D_{KL}(P_{true}(r|c,q) || P_\theta(r|c,q)) | (c,q,A) \in D_I] > E[D_{KL}(P_{true}(r|c,q) || P_\theta(r|c,q)) | (c,q,A) \in D_U]
    \end{equation}

    \item \textbf{Exploitation of Model Weaknesses:}
    \begin{equation}
    P(\text{Weakness Addressed} | (c,q,A) \in D_I) > P(\text{Weakness Addressed} | (c,q,A) \in D_U)
    \end{equation}
\end{enumerate}

This formalization demonstrates that training on informative datapoints, as generated by the GAP approach, leads to superior improvements in model performance across the entire input space, not just on the specific datapoints used for training.

\paragraph{Model Formulation}

Let $P(S_{ijk})$ be the probability that question $k$ submitted by player $i$ for image $j$ successfully reveals an AI mistake:

\begin{equation}
P(S_{ijk}) = \sigma(\alpha A_i - \beta D_j + \tau(1 - t_{ijk}/T) - \gamma F_{ij})
\end{equation}

where:
\begin{itemize}
    \item $\sigma(x) = \frac{1}{1 + e^{-x}}$ is the logistic function, mapping $\mathbb{R} \to (0,1)$
    \item $A_i \in [0,1]$ is the ability score of player $i$
    \item $D_j \in [0,1]$ is the difficulty score of image $j$
    \item $t_{ijk} \in [0,T]$ is the time taken by player $i$ to ask question $k$ for image $j$
    \item $T > 0$ is the total allowed time per image (e.g., 120 seconds)
    \item $F_{ij} \in [0,1]$ is the fatigue factor for player $i$ when reaching image $j$
    \item $\alpha, \beta, \tau, \gamma > 0$ are scaling parameters
\end{itemize}

\paragraph{Parameter Explanation and Domain Reasoning}

\begin{itemize}
    \item $A_i \in [0,1]$: Player ability score. Higher values indicate greater skill in identifying AI mistakes. The [0,1] range allows for intuitive interpretation and comparison between players.
    
    \item $D_j \in [0,1]$: Image difficulty score. Higher values represent more challenging images. The [0,1] range facilitates comparison across images and allows for intuitive difficulty scaling.
    
    \item $\alpha, \beta > 0$: Scaling parameters for player ability and image difficulty, respectively. These allow the model to adjust the relative importance of these factors.
    
    \item $t_{ijk} \in [0,T]$: Time taken for each question. Bounded by 0 and the maximum allowed time T.
    
    \item $T > 0$: Total allowed time per image, a positive constant defining the time limit.
    
    \item $\tau > 0$: Time pressure scaling parameter. This global parameter adjusts the impact of time pressure across all players and images.
    
    \item $F_{ij} \in [0,1]$: Fatigue factor. We propose modeling fatigue as:
    
    \begin{equation}
    F_{ij} = 1 - e^{-\lambda (j-1) / N}
    \end{equation}
    
    where $\lambda > 0$ is a fatigue rate parameter and $N$ is the total number of images. This formulation ensures $F_{ij}$ starts at 0 and asymptotically approaches 1 as j increases, capturing the cumulative effect of fatigue.
    
    \item $\gamma > 0$: Fatigue scaling parameter, adjusting the overall impact of fatigue on performance.
\end{itemize}

The term $(\alpha A_i - \beta D_j)$ represents the core interaction between player ability and image difficulty. The time pressure term $\tau(1 - t_{ijk}/T)$ increases the success probability for quicker responses, while the fatigue factor $\gamma F_{ij}$ decreases it as the session progresses.

\paragraph{Model Fitting on Tainted Dataset}

We fit the model using a tainted dataset where the ground truth about AI mistakes is known. The process involves:

\begin{enumerate}
    \item Data Collection: For each question in the tainted dataset, we record the player identifier, image identifier, question identifier, success outcome (1 if the question revealed an AI mistake, 0 otherwise), and time taken.
    
    \item Parameter Estimation: We use maximum likelihood estimation to fit $A_i$, $D_j$, $\alpha$, $\beta$, $\tau$, $\gamma$, and $\lambda$. The log-likelihood function is:
    
    \begin{equation}
    LL = \sum_i \sum_j \sum_k [S_{ijk} \log(P(S_{ijk})) + (1 - S_{ijk}) \log(1 - P(S_{ijk}))]
    \end{equation}
    
    where $S_{ijk}$ is the observed outcome for question $k$ by player $i$ on image $j$.
    
    \item Model Validation: We employ cross-validation, ROC curve analysis, and calibration plots to assess the model's performance and predictive power.
\end{enumerate}

\paragraph{Application to Untainted Dataset}

For the untainted dataset, where the ground truth is unknown, we apply the fitted model as follows:

\begin{enumerate}
    \item Probability Estimation: For each submitted question, we estimate $P(S_{ijk})$ using the fitted model parameters.
    
    \item Question Selection: We select questions that meet a predetermined probability threshold $\theta$:
    
    \begin{equation} \label{eq:que_selection}
    Q_j = \{q_{ijk} | P(S_{ijk}) > \theta\}
    \end{equation}
    
    where $Q_j$ is the set of selected questions for image $j$. This threshold helps focus on the most promising questions.
\end{enumerate}

\paragraph{Model Dynamics and Refinement}

To maintain relevance as players improve and new images are introduced, we implement:

\begin{itemize}
    \item Dynamic Player Ability: $A_{i,new} = A_{i,old} + \eta (\text{observed\_performance} - \text{expected\_performance})$, where $\eta > 0$ is a learning rate parameter.
    
    \item Image Clustering: Grouping similar images to estimate difficulty scores for new, unseen images.
\end{itemize}

This adaptive framework enables ongoing evaluation of player performance, identification of potential AI mistakes, and continuous improvement of both the evaluation model and the AI system being tested.

\end{document}